\documentclass{article} 
\usepackage{iclr2024_conference,times}

\usepackage{amsmath,amsfonts,bm}

\def\eqref#1{equation~\ref{#1}}

\def\1{\bm{1}}

\DeclareMathAlphabet{\mathsfit}{\encodingdefault}{\sfdefault}{m}{sl}
\SetMathAlphabet{\mathsfit}{bold}{\encodingdefault}{\sfdefault}{bx}{n}

\usepackage{hyperref}
\usepackage{url}
\usepackage{booktabs} 
\usepackage{amsfonts}       
\usepackage{nicefrac}       
\usepackage{microtype}      
\usepackage{xcolor}         
\usepackage{graphicx}
\usepackage{caption}
\usepackage{subcaption}
\usepackage{floatrow}
\usepackage{blindtext}
\usepackage{algorithm}
\usepackage{algorithmic}

\title{Evolving Loss Functions for Specific Image Augmentation Techniques}

\author{Brandon Morgan \& Dean Hougen \\
Department of Computer Science\\
University of Oklahoma\\
Norman, Oklahoma, USA \\
\texttt{\{morganscottbrandon, hougen\}@ou.edu} \\
}

\iclrfinalcopy 
\begin{document}

\maketitle

\begin{abstract}
Previous work in Neural Loss Function Search (NLFS) has shown a lack of correlation between smaller surrogate functions and large convolutional neural networks with massive regularization. We expand upon this research by revealing another disparity that exists, correlation between different types of image augmentation techniques. We show that different loss functions can perform well on certain image augmentation techniques, while performing poorly on others. We exploit this disparity by performing an evolutionary search on five types of image augmentation techniques in the hopes of finding image augmentation specific loss functions. The best loss functions from each evolution were then taken and transferred to WideResNet-28-10 on CIFAR-10 and CIFAR-100 across each of the five image augmentation techniques. The best from that were then taken and evaluated by fine-tuning EfficientNetV2Small on the CARS, Oxford-Flowers, and Caltech datasets across each of the five image augmentation techniques. Multiple loss functions were found that outperformed cross-entropy across multiple experiments. In the end, we found a single loss function, which we called the \emph{inverse bessel logarithm} loss, that was able to outperform cross-entropy across the majority of experiments.
\end{abstract}

\section{Introduction}

Neural loss function search (NFLS) is the field of automated machine learning dedicated to finding loss functions better than cross entropy for machine learning and deep learning tasks. Much ground has been covered in this field \citep{MetaLearning,BAIKAL,LossTaylor,LossObjectDetection,LossLabelNoise,LossDistributionalShift,SoftmaxPersonLoss,AutoLoss,LearnSymbLossMeta}. NLFS has been applied to object detection \citep{LossObjectDetection}, image segmentation \citep{AutoLoss}, and person re-identification \citep{SoftmaxPersonLoss}. Recently, \citet{brandon} proposed a new search space and surrogate function for the realm of NLFS. Specifically, they evolved loss functions for large scale convolutional neural networks (CNNs). In their work, they were able to show a disparity of correlation between smaller surrogate functions and EfficientNetV2Small. To circumnavigate this problem, they searched for a surrogate function that correlated well with their selected large scale CNN. 

From this observation, we further investigate this lack of correlation. Specifically, we show that loss functions can vary greatly in performance with different amounts of image augmentation. We construct five different image augmentation techniques to evaluate the loss functions. We show that there exists a lack of correlation between different image augmentation techniques and loss functions. Given this disparity, we perform an evolutionary search on the CIFAR-10 dataset for each of the five image augmentation techniques in the hopes of finding image augmentation specific loss functions. The best three loss functions from each evolutionary run were taken and transferred to WideResNet-28-10 (36M parameters) \citep{WideResNet} across all image augmentation techniques on CIFAR-10 and CIFAR-100 \citep{CIFAR}. The best loss functions overall were then taken and evaluated on the CARS-196 \cite{CARS196}, Oxford-Flowers-102 \citep{Flowers102}, and Caltech-101 \citep{Caltech} datasets by fine-tuning EfficientNetV2Small (20.3M parameters) \cite{EffNetV2}. In the end, we found one loss function, which we call the \emph{inverse bessel logarithm} function, that was able to outperform cross entropy on the majority of experiments. 

Our summarized findings reveal four key insights: (1) we confirm the results shown reported by \citet{brandon}, by again revealing the disparity between surrogate functions and large scale CNNs; (2) we reveal another disparity existing between loss functions transferred over different image augmentation technique; (3) we reveal that loss functions can perform extremely well on the evolved dataset, while performing extremely poorly on other datasets; and (4) we found one loss function in particular, the \emph{inverse bessel logarithm} loss, which was able to outperform cross-entropy across the majority of experiments. 

\section{Related Work}

\textbf{Cross-Entropy:} Machine and deep learning algorithms and models are built off the back of optimization theory: the minimization of an \emph{error} function. From information theory, \emph{entropy}, \emph{Kullback-Leibler divergence}  (KL), and \emph{cross-entropy} emerged early on as common error metrics to measure the \emph{information}, or randomness, of either one or two probability distributions. In these instances, the goal is to find a set of weights for a model, where the model's output distribution, given an input, matches the target distribution for that said input. The full mathematical relationship between the metrics can be shown below in Equation~\ref{eqn:ce}, where $P$ and $Q$ are probability distributions, $CE(P, Q)$ is cross-entropy, $E$ is entropy, and $KL$ is KL. In machine learning, $P$ is the distribution of the model's output, and $Q$ is the target distribution. Therefore, KL divergence is the difference between the cross-entropy of the model's output distribution and target distribution, and the randomness of the model's output distribution. In practice, cross-entropy is selected more often than KL, as deep learning models calculate the \emph{loss} over a sampled mini-batch, rather than the population.

\begin{equation} \label{eqn:ce}
\begin{aligned}
    CE(P, Q) &= KL(P || Q) + E(P) \\
-\sum_i p(x_i)\ln(q(x_i) &= -\sum_i p(x_i)\ln(\frac{q(x_i)}{p(x_i)}) -\sum_i p(x_i)\ln(p(x_i)) \\
\end{aligned}
\end{equation}

\noindent \textbf{Convolutional Models:} Convolutional neural networks (CNNs) are deep learning models specifically designed for handle images, due to the spatial and temporal composition of the convolution operation. From the success of the \emph{residual network} (ResNet) \citep{ResNetV2}, the era of CNNs exploded during the early to mid 2010's. In this work, two such successors will be used, the \emph{wide residual network} (WideResNet), and the \emph{efficient net version two small} (EffNetV2Small). Both of these CNNs are state-of-the-art models commonly utilized in bench-marking new computer vision techniques \citep{autoaugment,populationBasedAut,RandAug,ConvNext}. \\

\noindent \textbf{Image Augmentation Techniques:} Machine and deep learning models are heavily prone to \emph{over-fitting}, a phenomenon that occurs when the error of the training dataset is small, but the error on a validation or test dataset is large. For image processing, image data augmentations are commonly applied to add \emph{regularization} by artificially increasing the dataset size. Simple data augmentations include vertically or horizontally shifting, flipping, or translating an image. From the advent of preventing co-adaptation of features within neural networks through \emph{dropout} \citep{dropout}, \emph{cutout} \citep{cutout} was proposed to drop random cutouts of pixels from the input image before being fed into the model. By dropping out random patches of pixels from the input image, the model must learn to classify an object using the remaining data. Although it was originally applied for regression problems, \emph{mixup} \citep{mixup} has been successfully applied to classification. Mixup creates artificial images and labels by taking a random linear interpolation between two images. As a result, the model's output distribution becomes less bimodal, and more uniform, which un-intuitively increases generalization \citep{label_smoothing}. Besides the simple data augmentations mentioned earlier, there also exist many others, such as shearing, inversion, equalize, solarize, posterize, contrast, sharpness, brightness, and color. Many works have been published in order to discover the best set of \emph{policies}, combinations of image augmentations, for specific datasets  \citep{autoaugment, populationBasedAut}. However, \emph{random augment} (RandAug) \citep{RandAug} has been regarded as the only policy capable of generalizing across datasets with extremely little need of fine-tuning or excess training \citep{EffNetV2}. Given $n$ and $m$, RandAug performs $n$ random augmentations from a list of possible augmentations, performing each one at a strength of $m$. Simple augmentations, cutout, mixup, and RandAug are all state-of-the-art augmentation techniques that are currently be used in practice for different situations. \\

\noindent \textbf{Loss Functions:} Besides image augmentations or model regularization, generalization can be encouraged during learning through the loss function. \citet{focal} proposed the \emph{Focal} loss for specifically handling class imbalances in object detection. \citet{BAIKAL} proposed one of the first automated methods of searching for loss functions specifically for image classification. Since then, many others have contributed to the field of NFLS \citep{MetaLearning,BAIKAL,LossTaylor,LossObjectDetection,LossLabelNoise,LossDistributionalShift,SoftmaxPersonLoss,AutoLoss,LearnSymbLossMeta}. \\

\noindent \textbf{Search Space:} Recently, \citet{brandon} revealed a disparity of correlation between smaller surrogate models, used during searching, and full-scale models with regularization, used during evaluation. In their work, they tested various surrogate models to assess correlation, finding little to no rank correlation except when training for a shortened period of epochs on the full-scale model. Due to their success in evolving loss functions, and our expansion on their findings, we chose to utilize their proposed search space, genetic algorithm, and other components. In NLFS, the majority of works have used genetic programming, due to the tree-based grammar representations of their search spaces \citep{BAIKAL,LossObjectDetection, SoftmaxPersonLoss, AutoLoss, LearnSymbLossMeta}. However, \citet{brandon} created a derivative search space from the \emph{NASNet} neural architecture search space \citep{NASNet}. Due to the success of \emph{regularized evolution} \citep{AmoebaNet} on the original NASNet search space, it was the selected genetic algorithm for their  NASNet derivative loss search space. A basic and condensed overview of regularized evolution can be shown in Algorithm~\ref{alg:ga}. For specifics, please refer to the original work of \citet{brandon}. 

\begin{figure}[htb]
\vskip 0.1in
\begin{center}
\centerline{\includegraphics[width=0.5\columnwidth]{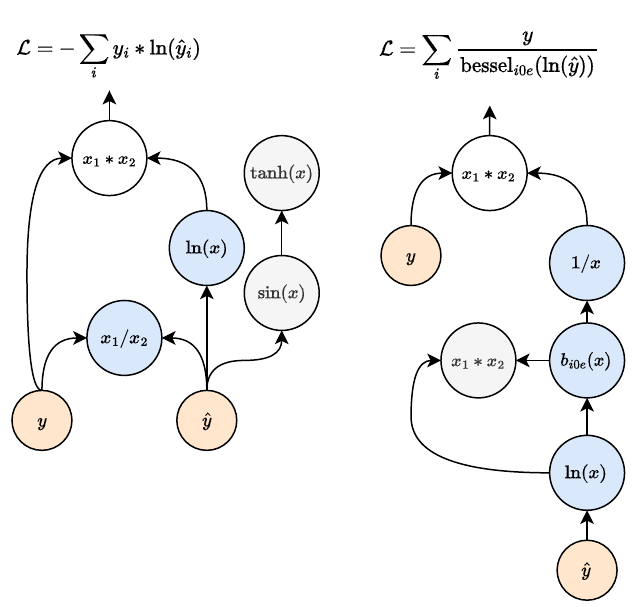}}
\caption{(Left) Example of how cross-entropy can be encoded using the search space. (Right) The exact encoding of the \emph{inverse bessel logarithm} function in the search space. The root node is in white, while the active nodes are in blue, inactive nodes are in grey, and the final loss equation is given above the root node.} 
\label{fig:ce_bessel_nodes}
\end{center}
\vskip -0.1in
\end{figure}

Their proposed search space is composed of free-floating nodes called \emph{hidden state nodes}. These nodes perform an operation on a receiving connection, and output the value to be used by subsequent nodes. The search space contains three components: (1) \emph{input connections}, which are analogous to leaf nodes for grammar based trees: $y$ and $\hat y$, where $y$ is the target output and $\hat y$ is the model's prediction; (2) \emph{hidden state nodes}, which are free-floating nodes that use any other hidden state nodes output, or input connection, as an argument; and (3) a specialized hidden state node, called the \emph{root} node, that is the final operation before output the loss value. Each hidden state node perform either a binary or unary operation. Their proposed 27 binary unary operations are listed in Table~\ref{tab:unary}. The seven proposed binary operations are available in Table~\ref{tab:binary}. Figure~\ref{fig:ce_bessel_nodes} shows how cross-entropy can be encoded in the search space, while also showing how the \emph{inverse bessel logarithm} function was encoded in the search space. 

\begin{table}[t]
\caption{The set of unary operations.}
\label{tab:unary}
\vskip 0.1in
\begin{center}
\begin{small}
\begin{tabular}{ll}
\toprule
\begin{sc}
Unary Operations
\end{sc}
 \\
\midrule
$-x$ & $\ln(|x|+\epsilon)$ \\
$e^x$ & $\log_{10}(|x|+\epsilon)$  \\
$\sigma(x)$ & $\frac{d}{dx} \sigma(x)$ \\
$\text{softsign}(x)$ & $\frac{d}{dx}\text{softsign}(x)$  \\
$\text{softplus}(x)$ & $\text{erf}(x)$ \\
$\text{erfc}(x)$ &  $\sin(x)$ \\
$\sinh(x)$ & $\text{arcsinh}(x)$ \\
$\tanh(x)$ & $\frac{d}{dx} \tanh(x)$ \\
$\text{arctanh}(x)$ &  $1/(x+\epsilon)$ \\
$|x|$ & $x^2$ \\
$\text{bessel}_{i0}(x)$ & $\sqrt{x}$ \\
$\text{bessel}_{i1}(x)$ & $\max(x, 0)$ \\
$\text{bessel}_{i1e}(x)$ & $\min(x, 0)$ \\
$\text{bessel}_{i0e}(x)$  & \\
\bottomrule
\end{tabular}
\end{small}
\end{center}
\vskip -0.1in
\end{table}

\begin{table}[t]
\caption{The set of binary operations.}
\label{tab:binary}
\vskip 0.1in
\begin{center}
\begin{small}
\begin{sc}
\begin{tabular}{ll}
\toprule
Binary Operations \\
\midrule
$x_1+x_2$ & $x_1*x_2$ \\
$x_1-x_2$ & $x_1/(x_2+\epsilon)$ \\
$x_1/\sqrt{1+x_2^2}$ & $\max(x_1, x_2)$ \\
$\min(x_1, x_2)$ & \\
\bottomrule
\end{tabular}
\end{sc}
\end{small}
\end{center}
\vskip -0.1in
\end{table}

\begin{algorithm}[htb]
   \caption{Regularized Evolution}
   \label{alg:ga}
\begin{algorithmic}[1]
    \STATE $population$ =  RandomInitialization()
    \STATE $validationAcc$ = TrainModel($population$)
    \REPEAT
    \STATE $winner$ = Tournament($population$, $validationAcc$)
    \STATE $child$ = Mutation($winner$)
    \STATE $valAcc$ = TrainModel($child$)
    \STATE $population$ = ReplaceOldest($population$, $child$)
    \STATE $validationAcc$ = Update($validationAcc$, $valAcc$)
    \UNTIL{Time Expired}
\end{algorithmic}
\end{algorithm}

\section{Methodology}

\subsection{Image Augmentation Correlation}
\label{sec:image_aug}

Because training each candidate loss function on a full scale model is extremely computationally expensive, \emph{surrogate} functions can be used as proxies. It is typical to use smaller models, smaller datasets, and shortened training schedules when constructed surrogate functions \citep{NASNet, ACT1, ACT2}. \citet{brandon} revealed the disparity of correlation between loss functions trained on smaller models and those trained on full scale models. Now, we investigate this observation further by taking a look at image augmentation techniques.  

\begin{figure*}[htb]
\vskip 0.1in
\begin{center}
\centerline{\includegraphics[width=\columnwidth]{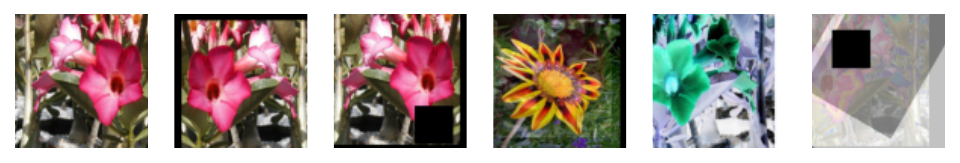}}
\caption{First Image: original untampered image. Second: base augmentation. Third: cutout augmentation. Fourth: mixup augmentation. Fifth: RandAug augmentation. Sixth: all augmentation} 
\label{fig:flower_aug}
\end{center}
\vskip -0.1in
\end{figure*}

In order to assess how well loss functions transferred across image augmentation techniques, we constructed five different image augmentations. The first, referred to as \emph{base}, served as the simplest form of image augmentation. In this augmentation, the images were zero-padded before a random crop equal to the training image size was obtained. The resulting image was then randomly horizontally flipped. The rest of the four image augmentations were built using \emph{base} as the foundation. The second, referred to as \emph{cutout}, applied the cutout operation, with a fixed patch size, to the image. The third, referred to as \emph{mixup}, applied the mixup operation by creating artificial images through random linear combinations of images and their labels. The fourth, referred to as \emph{RandAug}, applied the random augmentation strategy to randomly augment the images from a list of possible augmentations. Lastly, the fifth, referred to as \emph{all}, combined all previous augmentations together in order to maximize regularization. Figure~\ref{fig:flower_aug} shows an example of each augmentation on an image from the Oxford-Flowers dataset. As one can see, \emph{base} only slightly alters the image, while \emph{all} completely distorts it. Despite the massive distortion of the \emph{all} augmentation, our results reveal that the \emph{all} augmentation technique consistently outperforms the other four.

As with other neural component search methodologies, all searching experiments were performed on the CIFAR-10 dataset. For encouraging generalization, 5,000 of the 50,000 training images were set aside as validation. The validation accuracy is reported in all preliminary results. Using the search space, 1,000 random loss functions were generated and evaluated on the CIFAR-10 dataset. We did not have the computational resources to evaluate all loss functions on a state-of-the-art large-scale model; however, this does not mean we used a small model either. We chose to use a custom ResNet9v2 containing 6.5 million parameters as it was quick to train, while being relatively large (millions of parameters). The model was trained with the Adam optimizer \citep{Adam}, batch size of 128, and one cycle cosine decay learning rate schedule \citep{cycle} that warmed up the learning rate from zero to 0.001 after 1,000 steps, before being decayed back down to zero. The model was trained for 40 epochs, 500 steps per epoch, totalling to 20,000 steps.

\begin{figure}[htb]
\vskip 0.1in
\begin{center}
\centerline{\includegraphics[width=0.5\columnwidth]{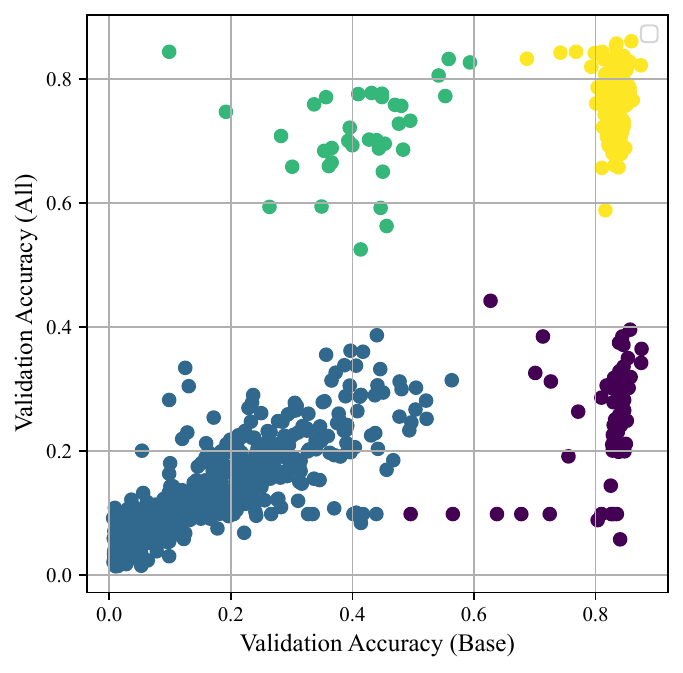}}
\caption{Validation accuracies for \emph{base} and \emph{all} techniques for the 1,000 randomly generated loss functions.} 
\label{fig:base_vs_all_corr_all}
\end{center}
\vskip -0.1in
\end{figure}

\begin{figure}[htb]
\vskip 0.1in
\begin{center}
\centerline{\includegraphics[width=0.5\columnwidth]{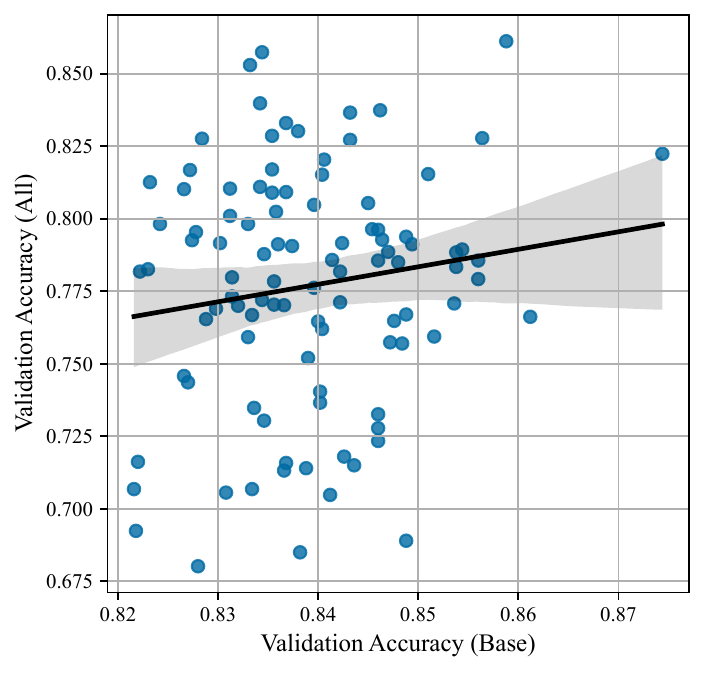}}
\caption{Validation accuracies for \emph{base} and \emph{all} techniques for the intersection of the best loss functions. } 
\label{fig:base_vs_all_corr_best}
\end{center}
\vskip -0.1in
\end{figure}

Figure~\ref{fig:base_vs_all_corr_all} shows the scatter-plot of the validation accuracies for all 1,000 loss functions evaluated on \emph{base} and \emph{all} image augmentations. As one can see, there appear to be four clusters of loss functions after running agglomerative clustering with average linkage (colors denote different clusters). The first contain losses that do not perform well on either \emph{base} ($<0.60$) nor \emph{all} ($<0.50$). The second contain losses that perform well on \emph{base} ($>0.80$) but not \emph{all} ($<0.50$). The third contain losses that perform decently well on \emph{all} ($>0.60$) but not well on \emph{base} ($<0.60$). The last contain losses that perform well on \emph{base} ($>0.75$) and decently well on \emph{all} ($>0.60$). Because we care about preserving the rank in-between loss functions on different image augmentation techniques, we rely upon Kendall's Tau rank correlation \citep{Kendall}. The top half of Table~\ref{tab:kendall_corr} gives the Tau rank correlation between all image augmentation techniques across all 1,000 randomly generated loss functions. As one can see, the rank correlations do not appear to poor, with the majority above $0.70$, signifying moderate to good rank preservation. However, Figure~\ref{fig:base_vs_all_corr_all} shows that the majority of loss functions generated are degenerate, meaning the loss functions perform poorly on base ($<0.80$), all ($<0.60$), or both. During evolution, these loss functions would get filtered out due to the early stopping mechanism of the surrogate function. When the intersection of best 50 loss functions from \emph{base} and \emph{all} are joined together, the rank disparity becomes much more clearer. Figure~\ref{fig:base_vs_all_corr_best} zooms in on the intersection of the best 50 loss functions from base and all. As one can see, the correlation seems almost uniform. The bottom half of Table~\ref{tab:kendall_corr} gives the Tau rank correlation between all image augmentation techniques across the intersection of the best 50 loss functions of each respective pair. These rank correlations are much lower, with all but one plummeting below $0.50$, indicating weak to moderate preservation of rank. Most noticeably, the rank correlations of \emph{base} between all other image augmentation techniques, except \emph{RandAug}, are extremely poor ($<0.10$). This disparity reveals that training loss functions with extremely little to no image augmentation (\emph{base}) have no guarantee to transfer well to other forms of image augmentation, even when using the same model, dataset, and training strategy. Perhaps these results could explain the observation first noted by \citep{brandon}, where their smaller models ($<0.3$M parameters), trained without any image augmentation, transferred poorly to EfficientNetV2Small, trained with \emph{RandAug}. Although, we cannot know this for certain without testing the effects of transferring loss functions across model size, this observation is important for all future NLFS endeavours, as it signifies that training loss functions with very little image augmentation may be predisposing them to lack the ability of effectively transferring to other augmentation techniques. 

\begin{table*}[htb]
\captionsetup{font=small}
\caption{Kendall Tau's rank correlation between the five different image augmentation techniques for 1000 randomly generated loss functions evaluated on a ResNet9 (6.5M parameters). Top half of the table is the rank correlation for all 1000 loss functions. Bottom half is the rank correlation for best 50.  }
\label{tab:kendall_corr}
\vskip 0.1in
\begin{center}
\begin{small}
\begin{sc}
\begin{tabular}{p{2cm} p{2cm} p{2cm} p{2cm}  p{2cm}}
\toprule
All Losses & Cutout & Mixup & RandAug & All \\
\midrule
\midrule
Base & 0.773 & 0.711 & 0.743 & 0.711   \\
Cutout  & - & 0.704 & 0.743 & 0.699  \\
Mixup  & & - & 0.692 & 0.736 \\
RandAug & & & - & 0.677 \\
\toprule
Best 50 Losses &  \\
\midrule
\midrule
Base  & 0.079 & 0.056 & 0.212 & 0.042   \\
Cutout  & - & 0.309 & 0.590 & 0.357  \\
Mixup  & & - & 0.442 & 0.462 \\
RandAug  & & & - & 0.482 \\
\bottomrule
\end{tabular}
\end{sc}
\end{small}
\end{center}
\vskip -0.1in
\end{table*}

\subsection{Evolutionary Runs}

From the results discussed in Section~\ref{sec:image_aug}, we decided to capitalize on the disparity of rank correlation between image augmentation techniques by performing regularized evolution on each of the five proposed augmentation techniques in the hopes of finding image augmentation specific loss functions. Five evolutionary runs were performed, one for each technique. We used the same setup for our evolution as described by \citet{brandon}. The best 100 loss functions from the initial 1,000 randomly generated loss functions formed the first generation for each respective technique. With a tournament size of 20, the genetic algorithm was allowed to run for 2,000 iterations. 

\subsection{Elimination Protocol}
\label{sec:elim_pro}
In order to properly evaluate the final loss functions, we chose to compare the final functions on WideResNet-28-10 (36M parameters), a large scale state-of-the-art model. To ensure transferability across model sizes, we used a similar loss function elimination protocol to \citet{brandon}. The best 100 loss functions discovered over the entire evolutionary search for each run were retrained on the same surrogate function as before, except the training session was extended to 192 epochs instead of 40, totalling to 96,000 steps. The best 24 from that were taken and trained on WideResNet-28-10 for each respective technique. The best 12 from that were taken and trained again. The best 6, compared by averaging their two previous runs on WideResNet, from that were taken and trained again. Finally, the best three, compared by averaging their previous three runs on WideResNet, were chosen and presented. When training on WideResNet-28-10, the SGD optimizer was used, along with Nesterov's momentum of 0.9, and a one cycle cosine decay learning rate schedule that warmed up the learning rate from zero to 0.1 after 1,000 steps, before being decayed back down to zero. The model was trained for 300 epochs, 391 steps per epoch, totalling to 117,300 steps.

\section{Results}

\subsection{Final Losses}

During our elimination protocol, the best 24 loss functions from the 100 trained on ResNet9v2 for 96,000 steps were trained on WideResNet for 117,300 steps. Using their final best validation accuracies for both models, we were able to construct the rank correlation to assess how well the loss functions would transfer to a larger model. Our results for \emph{base}, \emph{cutout}, \emph{mixup}, \emph{RandAug}, and \emph{all} were 0.243, 0.034, 0.112, 0.307, and 0.100. As one can see, these are extremely poor, which confirms the original findings by \citet{brandon} of poor correlation between smaller models and large CNNs. Despite this lack of correlation, the majority of our final found loss functions still performed well compared to CE. 

\begin{figure}[htb]
\vskip 0.1in
\begin{center}
\centerline{\includegraphics[width=0.5\columnwidth]{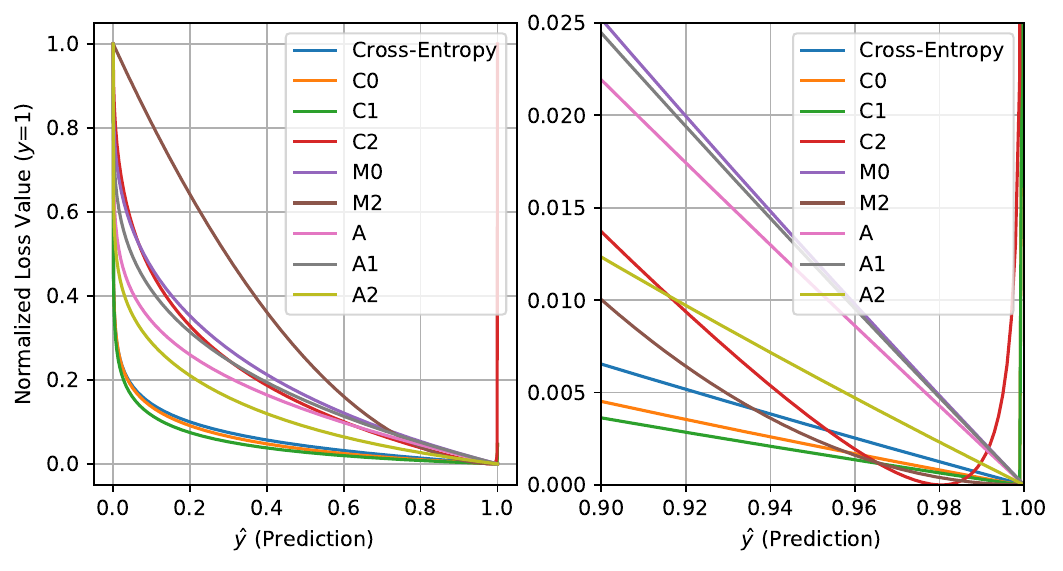}}
\caption{Binary phenotype of the final best loss functions against cross-entropy for normalized loss values. } 
\label{fig:best_losses}
\end{center}
\vskip -0.1in
\end{figure}

The final three loss functions for each augmentation technique, \emph{base} ($B0,B1,B2$), \emph{cutout} ($C0,C1,C2$), \emph{mixup} ($M0,M1,M2$), \emph{RandAug} ($R0,R1,R2$), and \emph{all} ($A0,A1,A2$), are shown in Table~\ref{tab:final_losses}. From the table, a few observations can be made. First, all \emph{base}, \emph{cutout}, and \emph{RandAug} functions used the expression $\hat y / (y + \epsilon)$. Second, all three of the \emph{all} functions were of the form $f(\log(\hat y))*y$, where it was either $\log(x)$ or $\ln(x)$, and $f(x)$ was dependent upon the function. Lastly, many of the loss functions, across all five of the techniques, used some form of the bessel function. The binary phenotypes (when $y=1$) of the normalized loss values for a hand selected number of the final loss functions, based upon the results in Section~\ref{sec:eval}, can be seen in Figure~\ref{fig:best_losses}. As one can see, many of the final loss functions differ in their phenotype compared to CE, either in terms of learning a shifted global minimum or a much steeper or less steep slope.

\begin{table}[t]
\caption{The final discovered loss functions.}
\label{tab:final_losses}
\vskip 0.1in
\begin{center}
\begin{small}
\begin{tabular}{p{0.75cm} p{6.5cm}}
\toprule
Name & Loss Equation \\
\midrule
B0 & $-\frac{1}{n}\sum_i\sigma(\text{softplus}(\hat y_i / (y_i + \epsilon)))$ \\
B1 & $-\frac{1}{n}\sum_i\frac{d}{dx}\sigma(\text{bessel\_i0e}(\hat y_i/ (y_i +\epsilon)))$ \\
B2 & $-\frac{1}{n}\sum_i\text{bessel\_i1e}(\text{erf}(\hat y_i / (y_i + \epsilon))$ \\
\midrule
C0 & $-\frac{1}{n}\sum_i\ln(\arctan(\hat y_i / (y_i + \epsilon)))$ \\
C1 & $-\frac{1}{n}\sum_i\ln(\text{softplus}(\hat y_i / (y_i + \epsilon))$ \\
C2 & $-\frac{1}{n}\sum_i\sigma(\log(\hat y_i / (y_i + \epsilon)))$ \\
\midrule
M0 & $-\frac{1}{n}\sum_i\text{bessel\_i1}(\arctan(\log(\hat y_i)))*y_i$ \\
M1 & $\frac{1}{n}\sum_i\ln(1+\exp(\hat y_i - y_i))$ \\
M2 & $\frac{1}{n}\sum_i\text{arcsinh}(\text{bessel\_i0}(y_i - \hat y_i))$ \\
\midrule
R0 & $\frac{1}{n}\sum_i\frac{d}{dx}\tanh(\text{bessel\_i0e}(\hat y_i / y_i + \epsilon)) +\text{bessel\_i0e}(\hat y_i / y_i)$ \\
R1 & $\frac{1}{n}\sum_i\log(\hat y_i / y_i+ \epsilon) / \sqrt{1 + (\log(\hat y_i / y_i+ \epsilon) + y_i)^2}$ \\
R2 & $\frac{1}{n}\sum_i|\log(\hat y_i / y_i+ \epsilon)|^{|\text{bessel\_i0}(y_i)|}$ \\
\midrule
A0 & $-\frac{1}{n}\sum_i\ln(\frac{d}{dx}\text{softsign}(\log(\hat y_i)))*y_i$ \\
A1 & $-\frac{1}{n}\sum_i\ln(\text{bessel\_i0e}(\log(\hat y_i))*y_i$ \\
A2 & $\frac{1}{n}\sum_i y_i / (\text{bessel\_i0e}(\ln(\hat y_i + \epsilon)))$ \\
\midrule
CE & $-\frac{1}{n}\sum_i y_i\ln\Bigl(|\hat y_i|+\epsilon\Bigr)$ \\
\bottomrule
\end{tabular}
\end{small}
\end{center}
\vskip -0.1in
\end{table}

\subsection{Evaluation}
\label{sec:eval}
To assess the quality of the final loss functions, all were trained on WideResNet-28-10 using the training strategy discussed in Section~\ref{sec:elim_pro} on the CIFAR-10 and CIFAR-100 datasets. The results for the CIFAR datasets can be seen in Table~\ref{tab:cifar}. For comparison, cross-entropy (CE), both with and without label smoothing (CE$_{0.10}$, $\alpha=0.10$) were trained. From the table, a few observations arise. First, for CIFAR-10, not all loss functions evolved for each specific image augmentation technique outperformed CE for that technique. For example, all of the final \emph{RandAug} loss functions performed poorly compared to CE on \emph{RandAug}. On the contrary, some loss functions performed well on techniques that they were not evolved upon. For example, functions from the \emph{all} group achieved Top~1 on \emph{base} and \emph{mixup}. Second, when transferring to CIFAR-100, all of the final functions for \emph{base} and \emph{RandAug}, and $M1$, collapse, performing extremely poorly. This observation reinforces the idea that loss functions evolved on a particular dataset, model, and training strategy may not perform well in other environments. Third, only twice did CE or CE$_{0.10}$ achieve Top~1 across all augmentation techniques on CIFAR-10 and CIFAR-100, indicating great success for the final loss functions. 

To further assess the performance of the final loss functions on different datasets and architectures, the best overall performing functions from CIFAR-10 and CIFAR-100 were selected be fine-tuned on CARS-196 using EfficientNetV2Small and their official ImageNet2012 \citep{ImageNet} published weights. The best overall functions were chosen to be all \emph{cutout} and \emph{all} functions, and $M0$ and $M2$. When fine-tuning on EfficientNetV2Small, the SGD optimizer was used, along with Nesterov's momentum of 0.9, batch size of 64, and a one cycle cosine decay learning rate schedule that warmed up the learning rate from zero to 0.1 after 1,000 steps, before being decayed back down to zero. The models were trained for 20 epochs, 500 steps per epoch, totalling to 10,000 steps. The results are shown in Table~\ref{tab:fine_results}. Once again, a few loss functions struggle to generalize, $C0$, $C2$, and $M2$. However, the $A2$ loss function performs relatively well compared to other discovered loss functions across all techniques. In fact, even combining $A2$ with a label smoothing value of $\alpha=0.10$, $A2_{0.10}$, achieves Top~1 for all techniques on CARS-196. The generalization power of $A2$ was further explored by being fine-tuned on Oxford-Flowers-102 and Caltech-101, as shown in Table~\ref{tab:fine_results}. For these other two datasets, the best loss flipped between $A2_{0.10}$ and CE$_{0.10}$. 

\begin{table*}[htb]
\captionsetup{font=small}
\caption{Results for all final loss functions evaluated on the CIFAR datasets using each of the five selected image augmentation techniques. The mean and standard deviation of the test accuracy across 5 independent runs are reported. Red indicates Top~1 performance, while bold indicates Top~5 performance.}
\label{tab:cifar}
\vskip 0.1in
\begin{center}
\begin{small}
\begin{sc}
\begin{tabular}{p{1.5cm} p{2cm} p{2cm} p{2cm} p{2cm}  p{2cm}}
\toprule
Loss & Base & Cutout & Mixup & RandAug & All \\
\midrule
\multicolumn{6}{c}{CIFAR-10} \\
\midrule
CE &  94.116$\pm$0.238 &95.400$\pm$0.107&\textbf{96.616}$\pm$0.600&95.012$\pm$0.261& \textbf{96.258}$\pm$0.227\\
CE$_{0.10}$ & 94.080$\pm$0.245 & 95.326$\pm$0.240 & 96.598$\pm$0.576 & 94.864$\pm$0.199 & \textbf{96.368}$\pm$0.244\\
\midrule
b0& 94.018$\pm$0.371 &95.222$\pm$0.106&94.396$\pm$0.358&94.510$\pm$0.153& 91.040$\pm$0.182\\
b1 &94.272$\pm$0.175  &95.324$\pm$0.066&95.212$\pm$0.397&94.628$\pm$0.226& 92.400$\pm$0.110\\
b2 &94.392$\pm$0.226  &95.544$\pm$0.090&95.552$\pm$0.531&94.902$\pm$0.116& 93.954$\pm$0.138\\
\midrule
c0 &\textbf{94.446}$\pm$0.202&\textbf{95.600}$\pm$0.154&95.934$\pm$0.441&95.022$\pm$0.169& 95.076$\pm$0.148\\
c1 &94.238$\pm$0.344 &95.496$\pm$0.048&95.936$\pm$0.634&\textbf{95.090}$\pm$0.276& 95.012$\pm$0.174\\
c2 &\textbf{94.454}$\pm$0.212 &\textcolor{red}{\textbf{95.710}}$\pm$0.141&95.030$\pm$0.482&\textcolor{red}{\textbf{95.164}}$\pm$0.143& 93.284$\pm$0.181\\
\midrule
m0&\textbf{94.432}$\pm$0.286&95.482$\pm$0.188&96.174$\pm$0.358&94.740$\pm$0.240& 96.044$\pm$0.125\\
m1&\textbf{94.446}$\pm$0.141&\textbf{95.528}$\pm$0.097&96.162$\pm$0.435&\textbf{95.092}$\pm$0.202& 95.268$\pm$0.080\\
m2 &94.364$\pm$0.322&\textbf{95.510}$\pm$0.065&\textbf{96.644}$\pm$0.447&\textbf{95.068}$\pm$0.233& 95.448$\pm$0.072\\
\midrule
r0 &94.250$\pm$0.267&95.434$\pm$0.165&93.136$\pm$0.404&94.660$\pm$0.227& 89.364$\pm$0.253\\
r1& 94.106$\pm$0.222&\textbf{95.514}$\pm$0.125&95.546$\pm$0.572&94.876$\pm$0.210& 94.204$\pm$0.046\\
r2 &93.704$\pm$0.344&95.172$\pm$0.184&95.486$\pm$0.468&93.858$\pm$0.263& 92.550$\pm$0.123\\
\midrule
a0 &94.258$\pm$0.140&95.488$\pm$0.147&\textcolor{red}{\textbf{96.654}}$\pm$0.507&94.758$\pm$0.270& \textbf{96.152}$\pm$0.287\\
a1 &\textcolor{red}{\textbf{94.566}}$\pm$0.178&95.390$\pm$0.130&96.086$\pm$0.081&94.866$\pm$0.249& \textbf{96.110}$\pm$0.185\\
a2 &94.304$\pm$0.231&95.374$\pm$0.149&\textbf{96.246}$\pm$0.440&\textbf{95.142}$\pm$0.224 & 96.076$\pm$0.239 \\
a2$_{0.10}$ & 93.946$\pm$0.276 & 95.270$\pm$0.100 & \textbf{96.644}$\pm$0.519 & 94.898$\pm$0.243 & \textcolor{red}{\textbf{96.418}}$\pm$0.19 \\
\midrule
\multicolumn{6}{c}{CIFAR-100} \\
\midrule
CE & 75.678$\pm$0.452 & \textbf{77.540}$\pm$0.472 & \textbf{80.054}$\pm$1.734 & \textcolor{red}{\textbf{77.354}}$\pm$0.448 & \textbf{81.166}$\pm$1.203  \\
CE$_{0.10}$ & \textbf{75.916}$\pm$0.391 & 76.692$\pm$0.703 & \textcolor{red}{\textbf{80.488}}$\pm$1.644& \textbf{76.522}$\pm$0.871 & \textbf{80.748}$\pm$1.666\\
\midrule
b0 & 59.652$\pm$1.505 & 41.734$\pm$0.602 & 45.960$\pm$0.833 & 31.988$\pm$0.425 & 31.988$\pm$0.425 \\
b1 & 21.852$\pm$0.893 & 20.504$\pm$1.966 & 31.602$\pm$1.071 & 11.940$\pm$1.665 &  16.392$\pm$0.448 \\
b2 & 54.822$\pm$2.624 & 22.892$\pm$2.199 & 78.464$\pm$1.162 & 22.892$\pm$2.199 & 49.318$\pm$1.225 \\
\midrule
c0 & 74.494$\pm$1.012  & 76.098$\pm$0.632 & \textbf{79.658}$\pm$2.038 & 75.654$\pm$0.661 & 79.412$\pm$0.640 \\
c1 & \textbf{75.748}$\pm$0.600 & 76.410$\pm$0.656 & 79.498$\pm$2.454 & 75.934$\pm$0.844 & 79.288$\pm$0.787 \\
c2 & \textcolor{red}{\textbf{76.694}}$\pm$0.419 & \textcolor{red}{\textbf{78.866}}$\pm$0.544 & 78.914$\pm$2.127 & \textbf{76.926}$\pm$0.614 & 72.340$\pm$0.545 \\
\midrule
m0 & 75.670$\pm$0.495 & \textbf{77.290}$\pm$0.626 & \textbf{79.670}$\pm$1.849 & 75.850$\pm$0.524 & 79.394$\pm$0.907 \\
m1 & 62.826$\pm$0.822 & 55.574$\pm$0.565 & 49.208$\pm$0.984 & 40.786$\pm$0.653 & 28.504$\pm$0.712 \\
m2 & \textbf{76.430}$\pm$0.585 & \textbf{77.278}$\pm$0.533 & 74.998$\pm$1.387 & 61.574$\pm$1.980 & 40.036$\pm$0.251 \\
\midrule
r0 & 31.658$\pm$0.676 & 26.438$\pm$0.455 & 25.345$\pm$0.881 & 15.038$\pm$0.706 & 10.280$\pm$0.451 \\
r1 & 17.732$\pm$1.834 & 30.760$\pm$1.598 & 46.834$\pm$1.729 & 15.120$\pm$0.117 & 12.291$\pm$0.374 \\
r2 & 73.854$\pm$0.666 & 60.856$\pm$0.859 & 51.300$\pm$1.668 & 25.148$\pm$2.357 & 15.964$\pm$0.383 \\
\midrule
a0 & 75.552$\pm$0.606 & 76.798$\pm$0.550 & 78.810$\pm$1.971 & 75.272$\pm$0.253 & \textbf{80.320}$\pm$1.475 \\
a1 & 75.278$\pm$0.589 & 76.630$\pm$0.727 & 78.394$\pm$1.946 & \textbf{76.652}$\pm$0.228 & \textbf{81.058}$\pm$0.814 \\
a2 & \textbf{76.034}$\pm$0.771 & \textbf{77.486}$\pm$0.460 & \textbf{80.252}$\pm$2.101 & \textbf{77.020}$\pm$0.729 & 79.378$\pm$0.427 \\
a2$_{0.10}$ & 75.714$\pm$0.609 & 76.954$\pm$0.482 & 79.114$\pm$1.853 & 76.516$\pm$0.688 & \textcolor{red}{\textbf{81.728}}$\pm$0.809 \\
\bottomrule
\end{tabular}
\end{sc}
\end{small}
\end{center}
\vskip -0.1in
\end{table*}

\begin{table*}[htb]
\captionsetup{font=small}
\caption{Results for all final loss functions evaluated when fine-tuning using each of the five selected image augmentation techniques. The mean and standard deviation of the test accuracy across 5 independent runs are reported. Red indicates Top~1 performance, while bold indicates Top~5 performance.}
\label{tab:fine_results}
\vskip 0.1in
\begin{center}
\begin{small}
\begin{sc}
\begin{tabular}{p{1.5cm} p{2cm} p{2cm} p{2cm} p{2cm}  p{2cm}}
\toprule
Loss & Base & Cutout & Mixup & RandAug & All \\
\midrule
\multicolumn{6}{c}{Cars-196} \\
\midrule

CE&88.740$\pm$0.195&89.506$\pm$0.124&88.390$\pm$0.189&\textbf{91.914}$\pm$0.077&91.747$\pm$0.076\\
CE$_{0.10}$&\textbf{89.628}$\pm$0.140&\textbf{89.643}$\pm$0.084&88.700$\pm$0.101&\textbf{91.909}$\pm$0.110&\textbf{91.765}$\pm$0.159\\
\midrule
c0&53.212$\pm$34.32&81.129$\pm$1.577&85.646$\pm$0.185&83.592$\pm$1.707&90.902$\pm$0.098\\
c1&\textbf{89.026}$\pm$0.893&\textbf{89.583}$\pm$0.112&85.654$\pm$0.368&90.593$\pm$2.959&90.969$\pm$0.203\\
c2&77.766$\pm$0.532&69.161$\pm$0.493&61.137$\pm$0.782&46.126$\pm$1.119&31.501$\pm$0.425\\
\midrule
m0&84.932$\pm$0.245&88.394$\pm$0.150&88.394$\pm$0.150&90.728$\pm$0.213&90.899$\pm$0.085\\
m2&72.459$\pm$1.016&36.426$\pm$1.270&36.426$\pm$1.270&58.866$\pm$1.864&11.800$\pm$1.122\\
\midrule
a0&86.758$\pm$0.214&88.638$\pm$0.186&\textbf{88.638}$\pm$0.186&91.063$\pm$0.152&91.466$\pm$0.123\\
a1&87.136$\pm$0.283&88.019$\pm$0.165&\textbf{88.780}$\pm$0.116&91.071$\pm$0.056&91.404$\pm$0.135\\
a2&88.820$\pm$0.190&89.255$\pm$0.196&88.720$\pm$0.157&91.899$\pm$0.077&\textbf{91.775}$\pm$0.108 \\
a2$_{0.10}$ &\textcolor{red}{\textbf{89.810}}$\pm$0.117&\textcolor{red}{\textbf{89.909}}$\pm$0.259&\textcolor{red}{\textbf{88.999}}$\pm$0.122&\textcolor{red}{\textbf{91.989}}$\pm$0.095&\textcolor{red}{\textbf{91.780}}$\pm$0.172 \\
\midrule
\multicolumn{6}{c}{Oxford-Flowers-102} \\
\midrule
CE & 89.403$\pm$0.338&89.985$\pm$0.477&87.985$\pm$0.426&93.283$\pm$0.205&92.181$\pm$0.171\\
CE$_{0.10}$ & \textcolor{red}{\textbf{92.613}}$\pm$0.110&\textcolor{red}{\textbf{93.710}}$\pm$0.415&\textcolor{red}{\textbf{89.816}}$\pm$0.422&94.786$\pm$0.156&92.753$\pm$0.211\\
A2 & 89.605$\pm$0.228& 89.855$\pm$0.663&88.769$\pm$0.331&93.644$\pm$0.333&92.340$\pm$0.273\\
A2$_{0.10}$ & 91.778$\pm$0.293&92.961$\pm$0.483&89.631$\pm$0.270&\textcolor{red}{\textbf{94.819}}$\pm$0.167&\textcolor{red}{\textbf{93.020}}$\pm$0.276\\
\midrule
\multicolumn{6}{c}{Caltech-101} \\
\midrule
CE & 90.621$\pm$0.629&91.318$\pm$0.411&90.053$\pm$0.203&90.595$\pm$0.308&91.085$\pm$0.137\\
CE$_{0.10}$ & \textcolor{red}{\textbf{93.261}}$\pm$0.241& \textcolor{red}{\textbf{92.932}}$\pm$0.451&90.756$\pm$0.076&\textcolor{red}{\textbf{91.437}}$\pm$0.235&91.381$\pm$0.158\\
A2 & 90.510$\pm$0.578&91.032$\pm$0.233&90.746$\pm$0.220&90.756$\pm$0.174&91.561$\pm$0.232\\
A2$_{0.10}$ & 93.041$\pm$0.234&92.794$\pm$0.397& \textcolor{red}{\textbf{90.986}}$\pm$0.394& 91.252$\pm$0.276 & \textcolor{red}{\textbf{91.423}}$\pm$0.319 \\
\bottomrule
\end{tabular}
\end{sc}
\end{small}
\end{center}
\vskip -0.1in
\end{table*}

\subsection{Inverse Bessel Logarithm Loss}

\begin{figure}[htb]
\vskip 0.1in
\begin{center}
\centerline{\includegraphics[width=0.5\columnwidth]{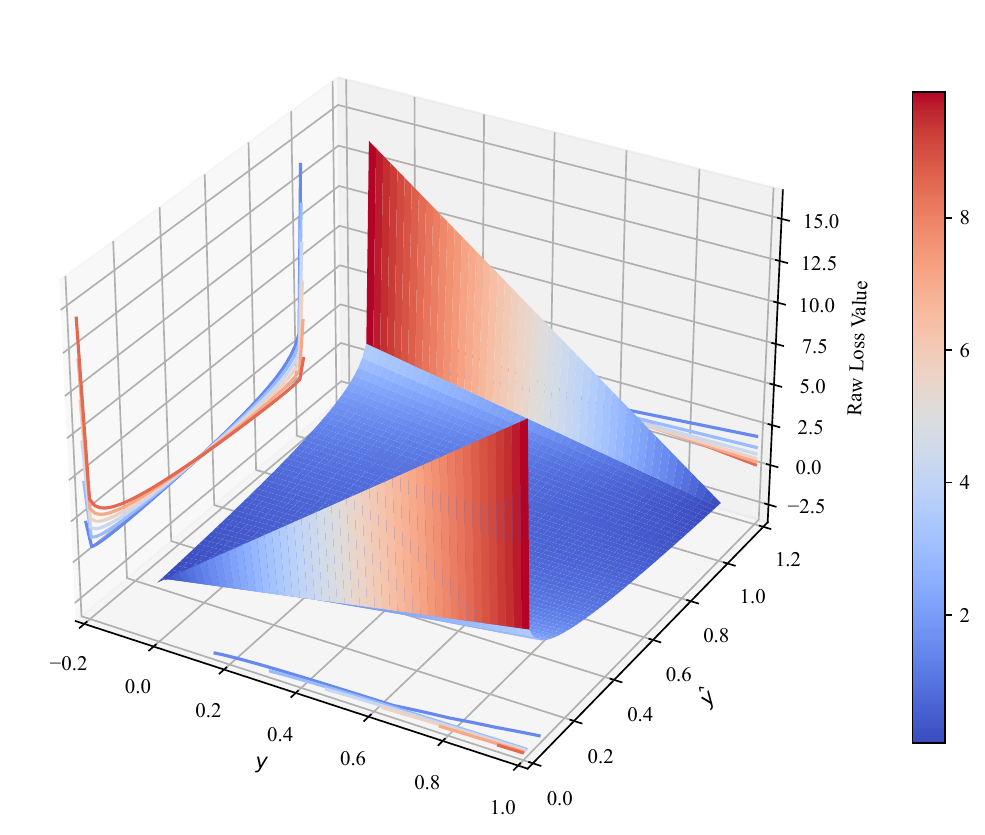}}
\caption{3D plot of the raw phenotype of the cross-entropy loss.} 
\label{fig:ce_3D}
\end{center}
\vskip -0.1in
\end{figure}

\begin{figure}[htb]
\vskip 0.1in
\begin{center}
\centerline{\includegraphics[width=0.5\columnwidth]{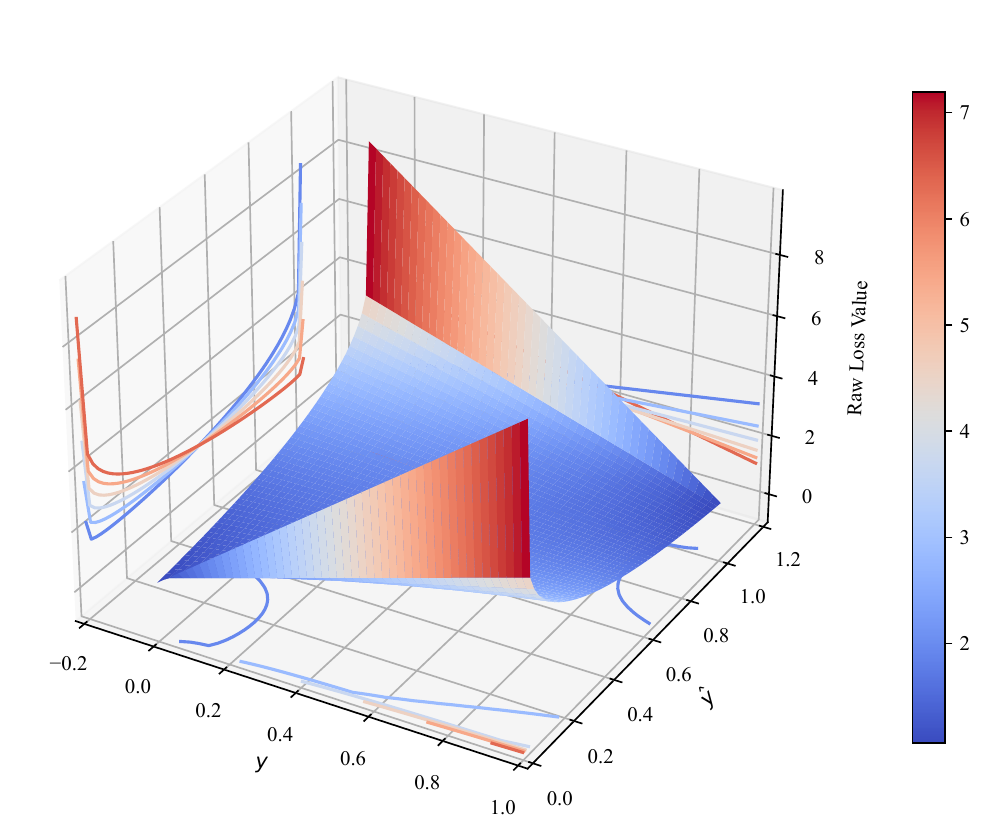}}
\caption{3D plot of the raw phenotype of the $A2$ loss.} 
\label{fig:a2_3D}
\end{center}
\vskip -0.1in
\end{figure}

\begin{figure}[htb]
\vskip 0.1in
\begin{center}
\centerline{\includegraphics[width=0.5\columnwidth]{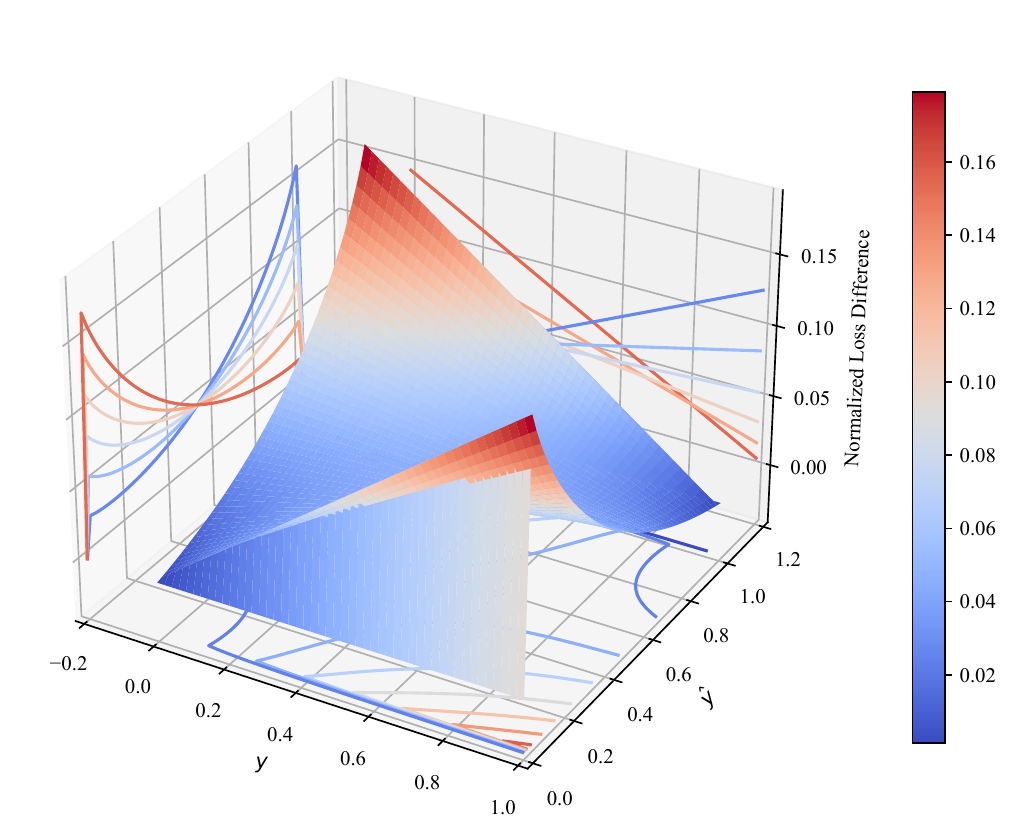}}
\caption{3D plot of the difference in normalized loss values between $A2$ and cross-entropy ($A2-CE$).} 
\label{fig:diff_3D}
\end{center}
\vskip -0.1in
\end{figure}

Due to the success of $A2$ in terms of generalization across the CIFAR and fine-tuning datasets, it was deemed as the final best loss function discovered in this work, which we called the \emph{inverse bessel logarithm loss}, due to its equation phenotype. The binary phenotypes of loss functions limit how well the loss landscape can be investigated for multiple dimensions. In order to better assess the differences between CE and the \emph{inverse bessel logarithm}, a three-dimensional plot can be constructed where the $x$-axis takes on $y$ values, the $y$-axis takes on $\hat y$ values, and the $z$-axis takes on the loss $\mathcal{L}(y, \hat y)$. Figure~\ref{fig:ce_3D} shows the non-normalized three-dimensional representation of CE, while Figure~\ref{fig:a2_3D} shows the non-normalized three-dimensional representation of \emph{inverse bessel logarithm}. Visually, one can see that \emph{inverse bessel logarithm} has a slight upward curve when $\hat y \sim 0$ and $\hat y \sim 1$. To better see this difference, Figure~\ref{fig:diff_3D} shows the normalized three-dimensional difference between \emph{inverse bessel logarithm} and CE. It can be seen that the largest differences occur right before $y$ and $\hat y$ are the most different (when $\hat y \to 0$ and $y \to 1$, or $\hat y \to 1$ and $y \to 0$). When $\hat y \sim 0$ and $y \sim 1$, the differences are extremely small. However, when $\hat y$ is slightly increased to $0.02$, the difference reaches its max value of $0.188$. As a result, it can be stated that \emph{inverse bessel logarithm} has larger normalized loss values than CE when the model is almost at its most incorrect state. This typically occurs only during the very early stages of training, when the model is at its most incorrect state. Perhaps these larger normalized loss values allow the model to expedite maneuvering to a better loss landscape early on during training, as larger loss values directly increase the learning rate.  

\section{Conclusion}

In this work, we further explored the results of recent work in NLFS that revealed a disparity of rank correlation of loss functions between smaller surrogate functions and large scale models with regularization. Specifically, we revealed another disparity that exists, the lack of rank correlation of loss functions between different image augmentation techniques. We showed that the best performing loss functions, from a set of randomly generated loss functions, across five different image augmentation techniques showed extremely poor rank correlation. We capitalized on this disparity by evolving loss functions on each respective augmentation technique in the hopes to find augmentation specific losses. In the end, we found a set of loss functions that outperformed cross-entropy across all image augmentation techniques, except one, on the CIFAR datasets. In addition, we found one loss function in particular ($A2$), which we call the \emph{inverse bessel logarithm}, that performed extremely well when fine-tuning on three large scale image resolution datasets. Our code can be found here: \url{https://github.com/OUStudent/EvolvingLossFunctionsImageAugmentation}. 

\section*{Acknowledgements}
The computing for this project was performed at the OU Super computing Center for Education and Research (OSCER) at the University of Oklahoma (OU).

\newpage

\bibliography{iclr2024_conference}
\bibliographystyle{iclr2024_conference}

\newpage

\appendix

\section{Binary and Three Dimensional Phenotypes}
\label{app:pheno}

In this appendix, the binary and three-dimensional phenotypes of the discovered loss functions will be covered.

\subsection{Base Losses}

\begin{figure}[htb]
\vskip 0.1in
\begin{center}
\centerline{\includegraphics[width=0.5\columnwidth]{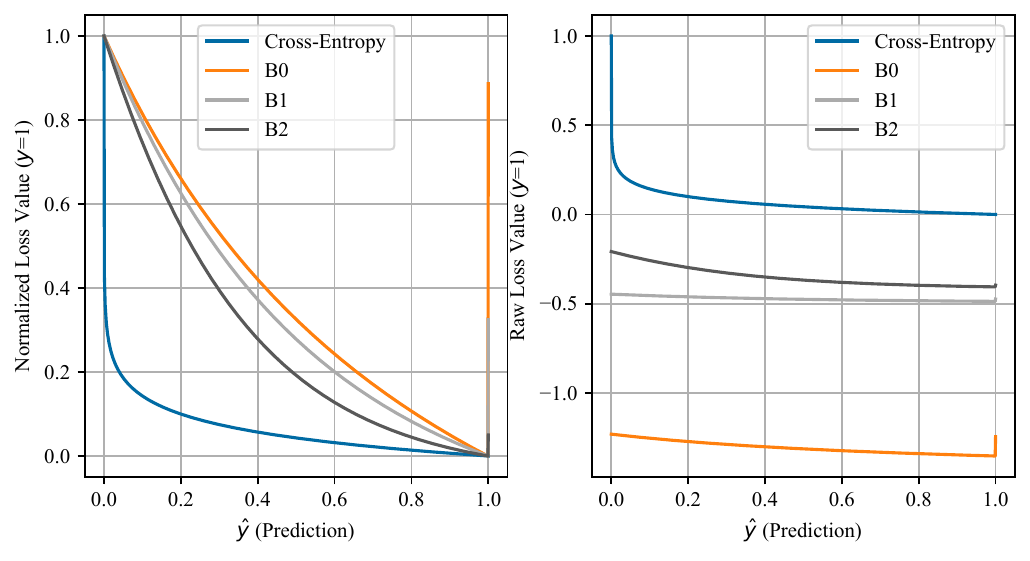}}
\caption{Binary Phenotype of the final Base losses against cross-entropy for both normalized and raw loss values.} 
\label{fig:base_losses}
\end{center}
\vskip -0.1in
\end{figure}

\begin{figure}[htb]
\vskip 0.1in
\begin{center}
\centerline{\includegraphics[width=0.5\columnwidth]{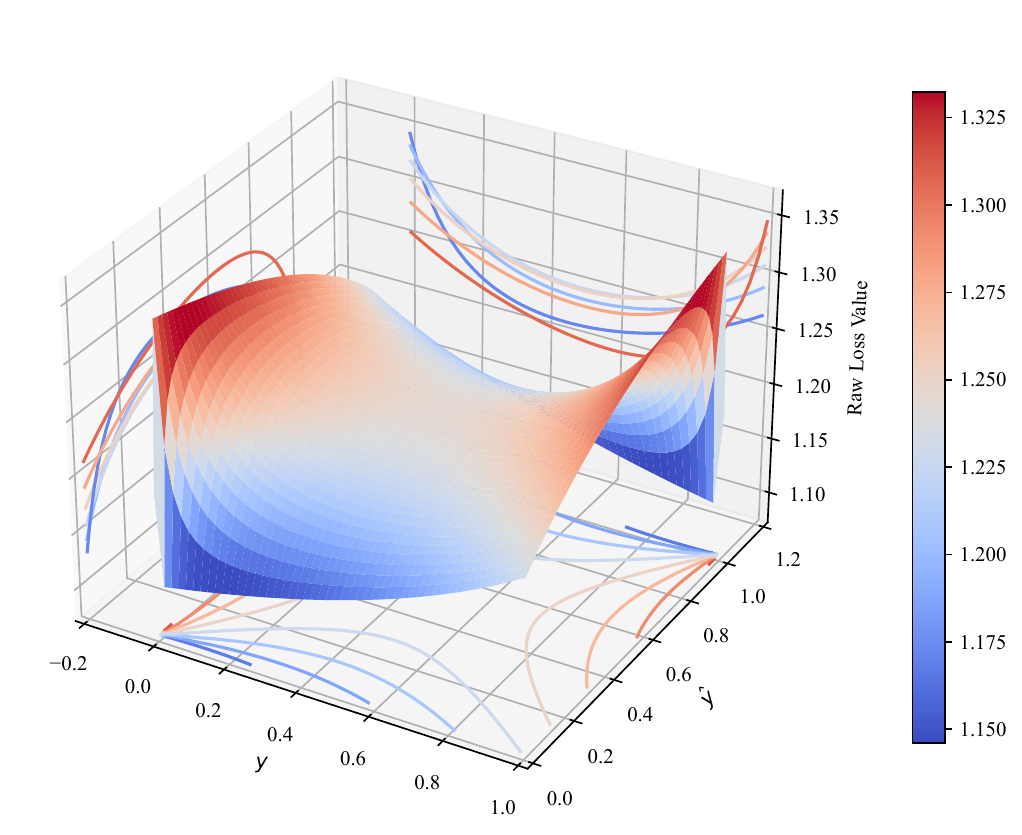}}
\caption{3D plot of the raw phenotype of the $B0$ loss.} 
\label{fig:b0}
\end{center}
\vskip -0.1in
\end{figure}

\begin{figure}[htb]
\vskip 0.1in
\begin{center}
\centerline{\includegraphics[width=0.5\columnwidth]{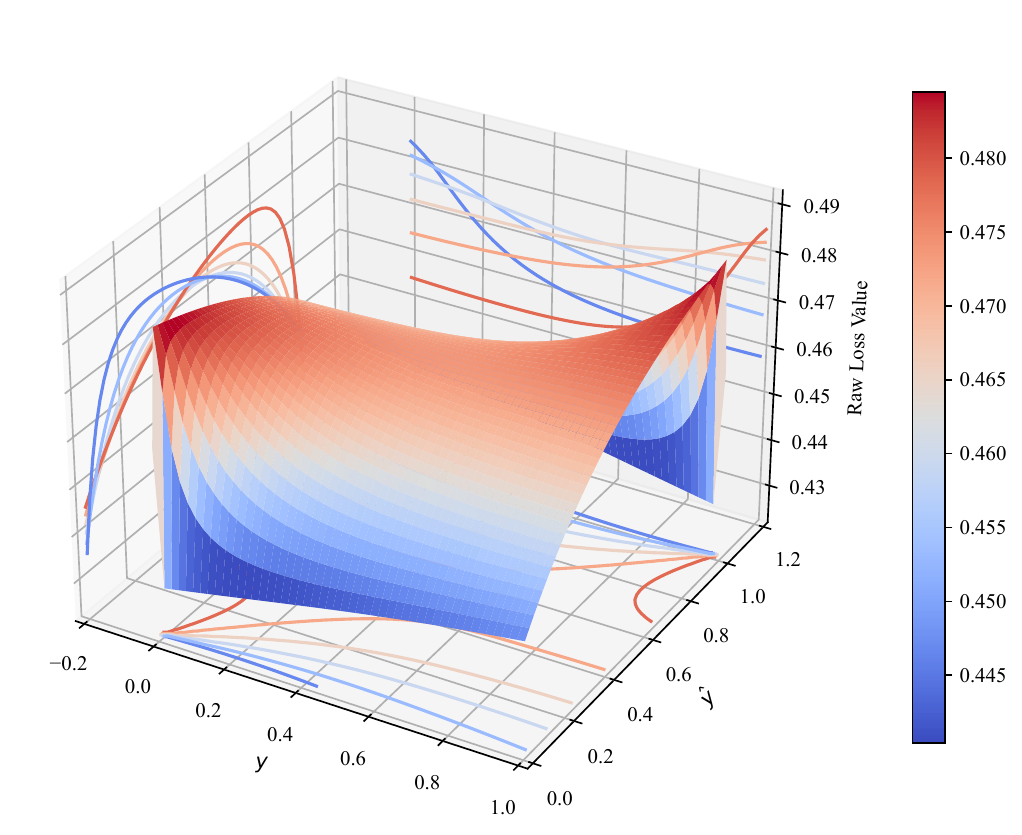}}
\caption{3D plot of the raw phenotype of the $B1$ loss.} 
\label{fig:b1}
\end{center}
\vskip -0.1in
\end{figure}

\begin{figure}[htb]
\vskip 0.1in
\begin{center}
\centerline{\includegraphics[width=0.5\columnwidth]{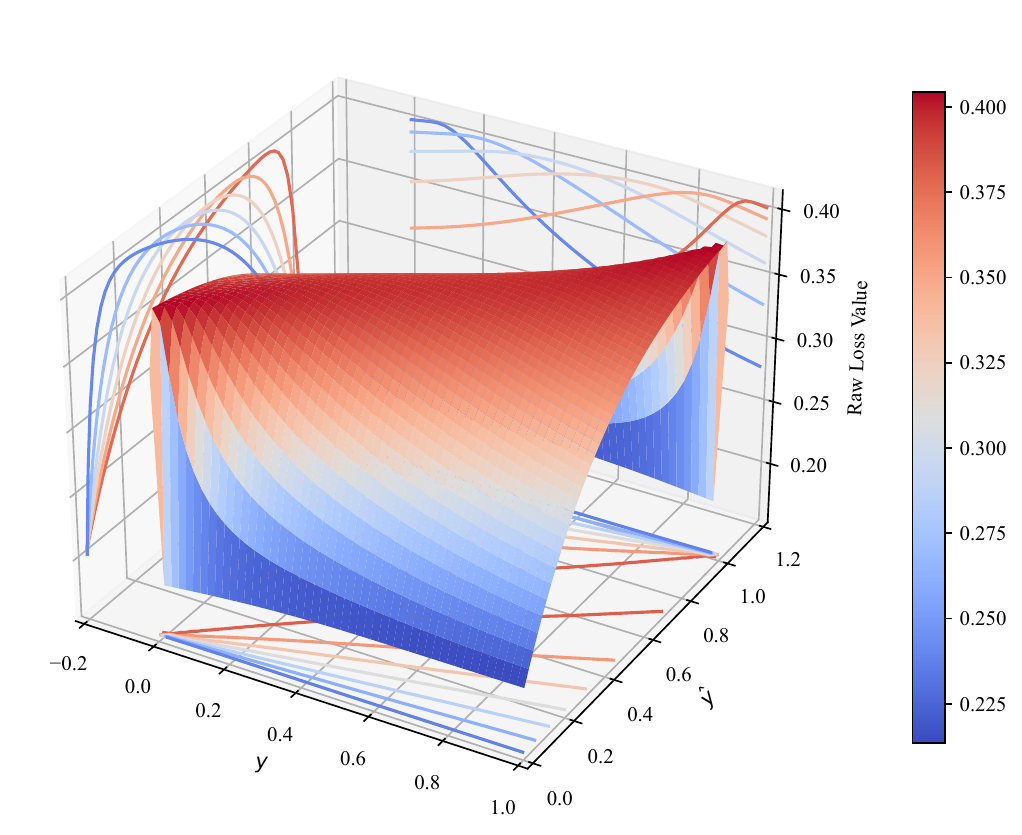}}
\caption{3D plot of the raw phenotype of the $B2$ loss.} 
\label{fig:b2}
\end{center}
\vskip -0.1in
\end{figure}

The normalized and raw binary phenotypes of the final base losses are shown in Figure~\ref{fig:base_losses}. The non-normalized three-dimensional representations for $B0$, $B1$, and $B2$ are shown in Figures~\ref{fig:b0}, \ref{fig:b1}, and \ref{fig:b2}. As one can see, the base functions are extremely different from CE. Specifically, their slope values are much less steeper than CE. Their three-dimensional representations look like saddles. Perhaps these saddle representations pose problems at higher dimensions, more than one class, as all base functions collapse when transferring to CIFAR-100. 

\subsection{Cutout Losses}

\begin{figure}[htb]
\vskip 0.1in
\begin{center}
\centerline{\includegraphics[width=0.5\columnwidth]{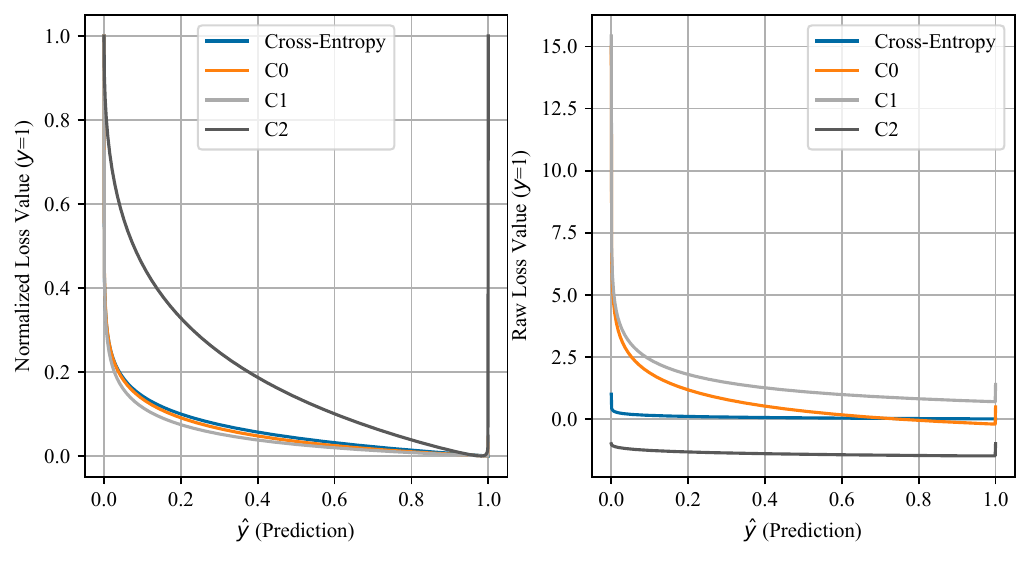}}
\caption{Binary Phenotype of the final Cutout losses against cross-entropy for both normalized and raw loss values.} 
\label{fig:cut_losses}
\end{center}
\vskip -0.1in
\end{figure}

\begin{figure}[htb]
\vskip 0.1in
\begin{center}
\centerline{\includegraphics[width=0.5\columnwidth]{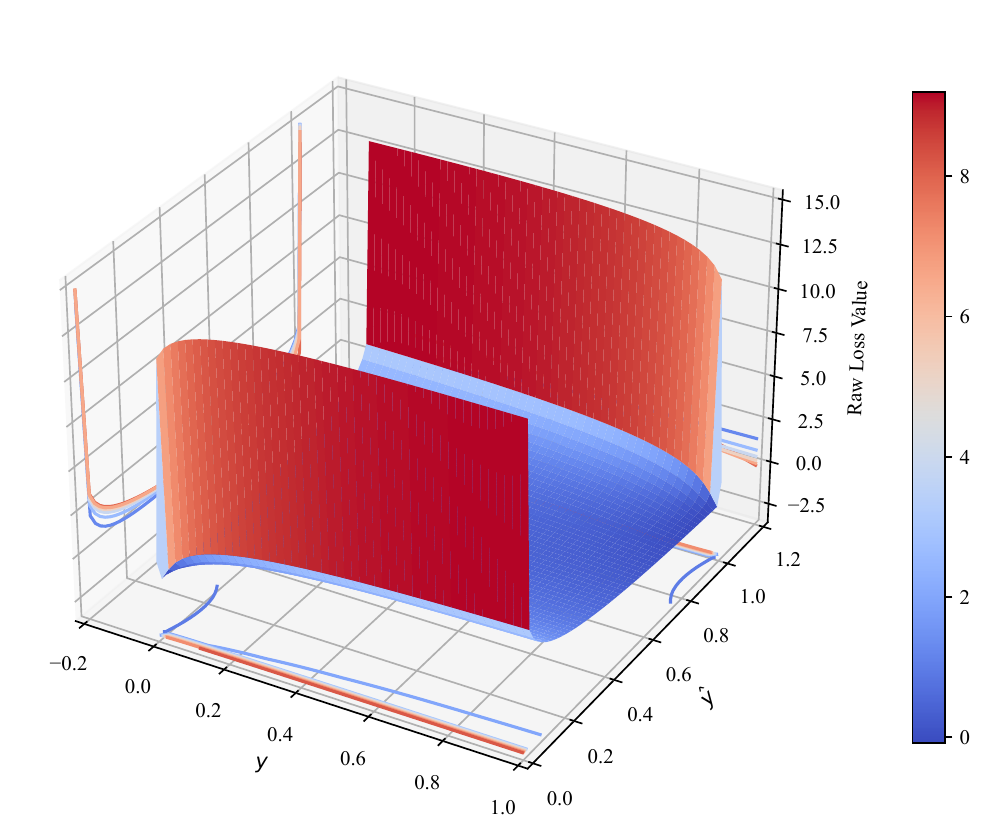}}
\caption{3D plot of the raw phenotype of the $C0$ loss.} 
\label{fig:c0_3D}
\end{center}
\vskip -0.1in
\end{figure}

\begin{figure}[htb]
\vskip 0.1in
\begin{center}
\centerline{\includegraphics[width=0.5\columnwidth]{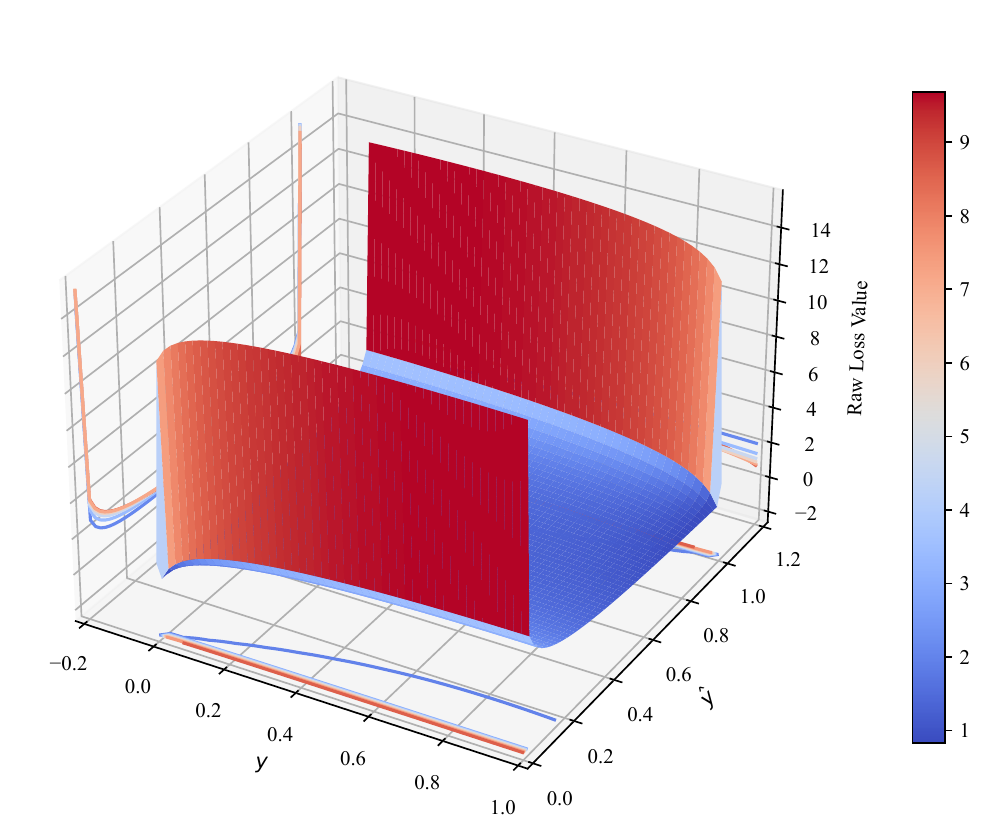}}
\caption{3D plot of the raw phenotype of the $C1$ loss.} 
\label{fig:c1_3D}
\end{center}
\vskip -0.1in
\end{figure}

\begin{figure}[htb]
\vskip 0.1in
\begin{center}
\centerline{\includegraphics[width=0.5\columnwidth]{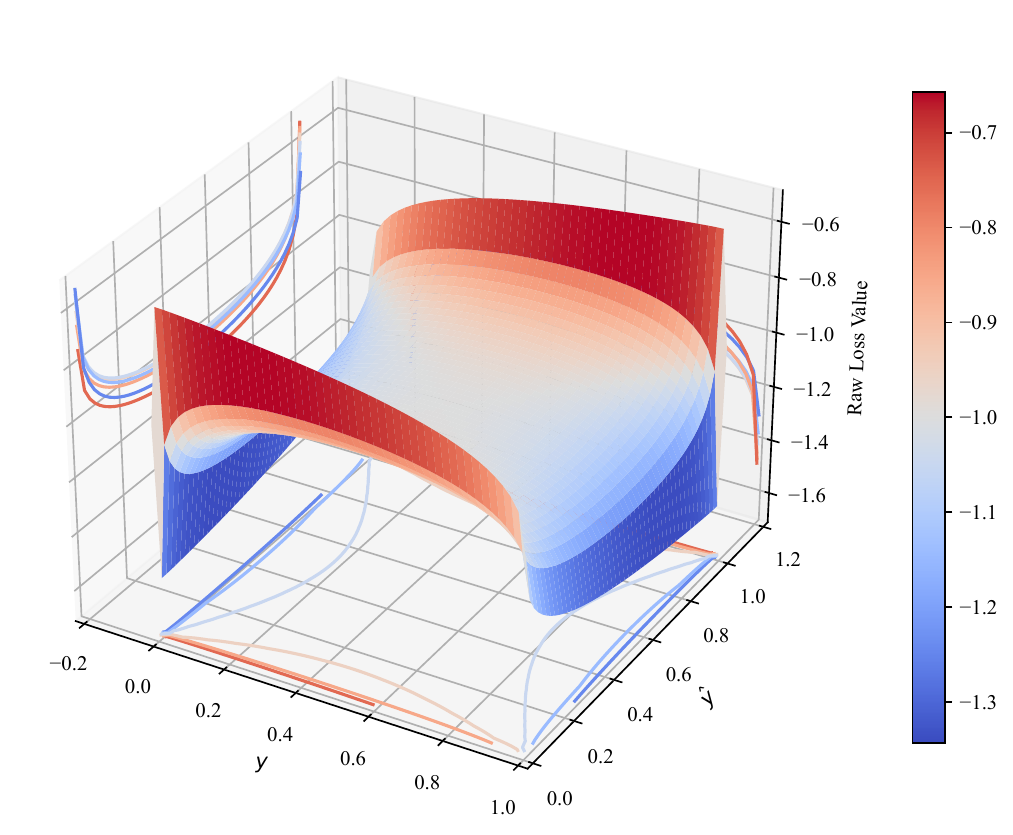}}
\caption{3D plot of the raw phenotype of the $C2$ loss.} 
\label{fig:c2_3D}
\end{center}
\vskip -0.1in
\end{figure}

The normalized and raw binary phenotypes of the final cutout losses are shown in Figure~\ref{fig:cut_losses}. The non-normalized three-dimensional representations for $C0$, $C1$, and $C2$ are shown in Figures~\ref{fig:c0_3D}, \ref{fig:c1_3D}, and \ref{fig:c2_3D}. As one can see from the binary phenotypes, the final cutout functions learn a little hook around $\hat y \sim 1$, a small label smoothing value. Unlike the base losses, the cutout losses are not so similar in their three-dimensional representations, despite having similar binary phenotypes. For example, $C0$ and $C1$ are similar, but are very different from $C2$. 

\subsection{Mixup Losses}

\begin{figure}[htb]
\vskip 0.1in
\begin{center}
\centerline{\includegraphics[width=0.5\columnwidth]{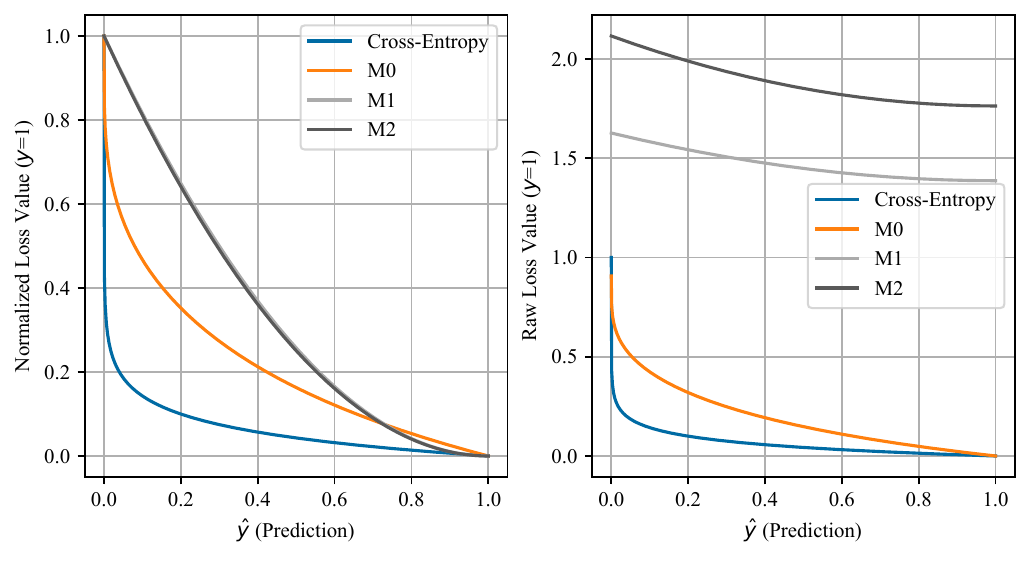}}
\caption{Binary Phenotype of the final Mixup losses against cross-entropy for both normalized and raw loss values.} 
\label{fig:mix_losses}
\end{center}
\vskip -0.1in
\end{figure}

\begin{figure}[htb]
\vskip 0.1in
\begin{center}
\centerline{\includegraphics[width=0.5\columnwidth]{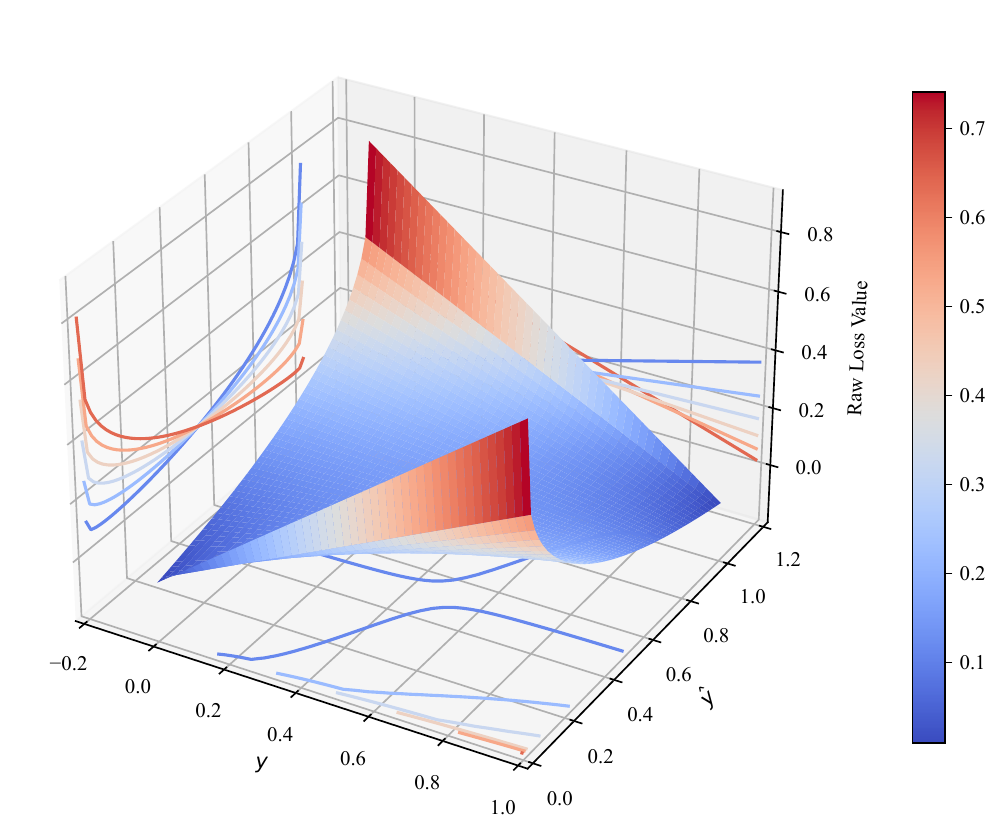}}
\caption{3D plot of the raw phenotype of the $M0$ loss.} 
\label{fig:m0_3D}
\end{center}
\vskip -0.1in
\end{figure}

\begin{figure}[htb]
\vskip 0.1in
\begin{center}
\centerline{\includegraphics[width=0.5\columnwidth]{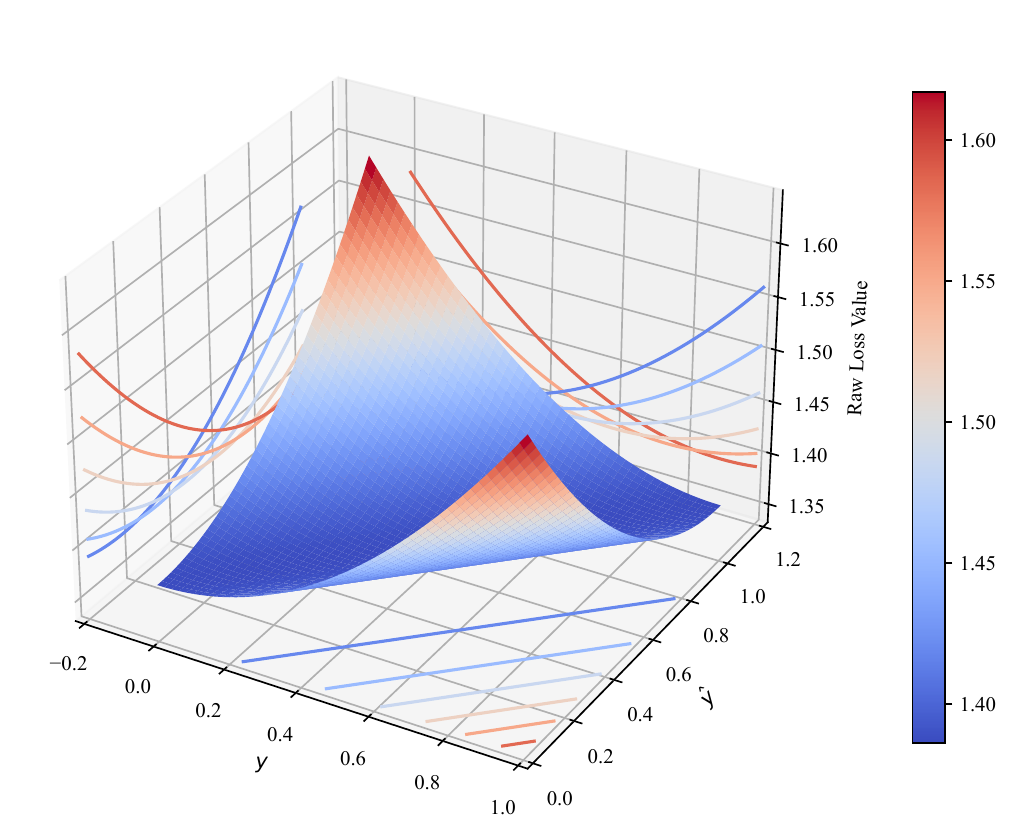}}
\caption{3D plot of the raw phenotype of the $M1$ loss.} 
\label{fig:m1_3D}
\end{center}
\vskip -0.1in
\end{figure}

\begin{figure}[htb]
\vskip 0.1in
\begin{center}
\centerline{\includegraphics[width=0.5\columnwidth]{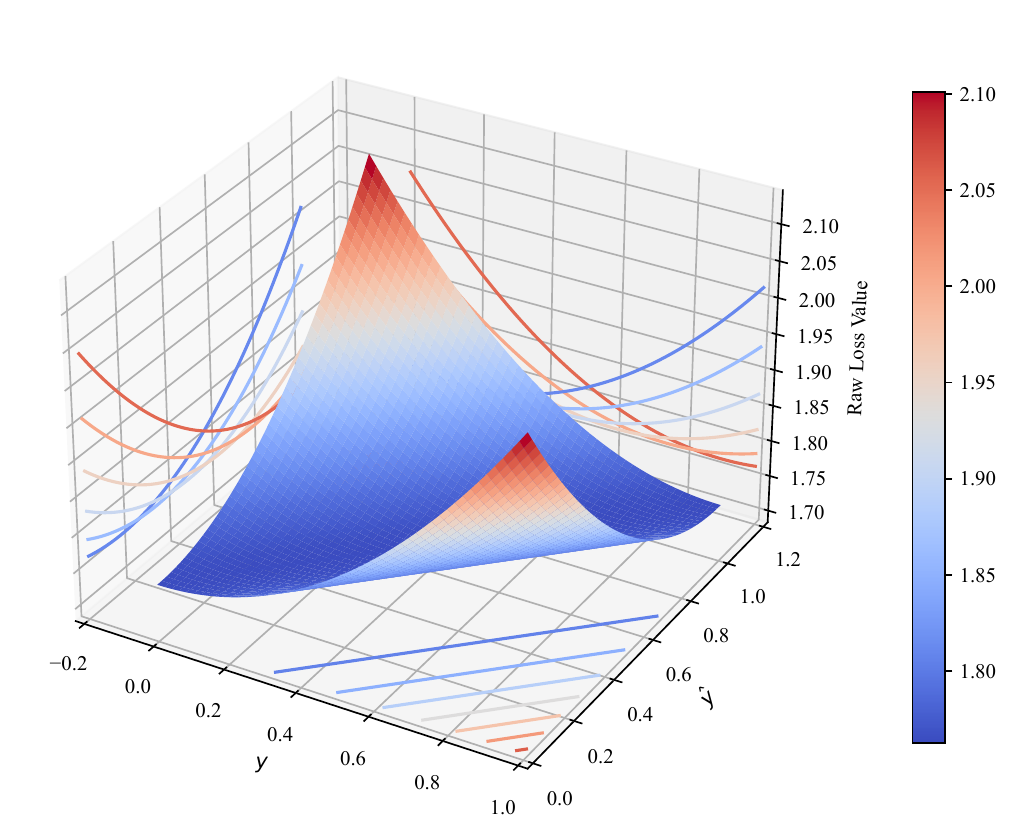}}
\caption{3D plot of the raw phenotype of the $M2$ loss.} 
\label{fig:m2_3D}
\end{center}
\vskip -0.1in
\end{figure}

The normalized and raw binary phenotypes of the final mixup losses are shown in Figure~\ref{fig:mix_losses}. The non-normalized three-dimensional representations for $M0$, $M1$, and $M2$ are shown in Figures~\ref{fig:m0_3D}, \ref{fig:m1_3D}, and \ref{fig:m2_3D}. As one can see from the binary phenotypes, the final mixup loss functions learn less steeper slopes than CE. In addition, $M2$ and $M1$ appear to have the same normalized values, indicating that they have the exact same shape, but different scaling. From the three-dimensional representations, $M0$ is different from $M1$ and $M2$, better resembling CE, except with more exaggerated curves when $y$ and $\hat y$ are most different, resembling $A2$.

\subsection{RandAug Losses}

\begin{figure}[htb]
\vskip 0.1in
\begin{center}
\centerline{\includegraphics[width=0.5\columnwidth]{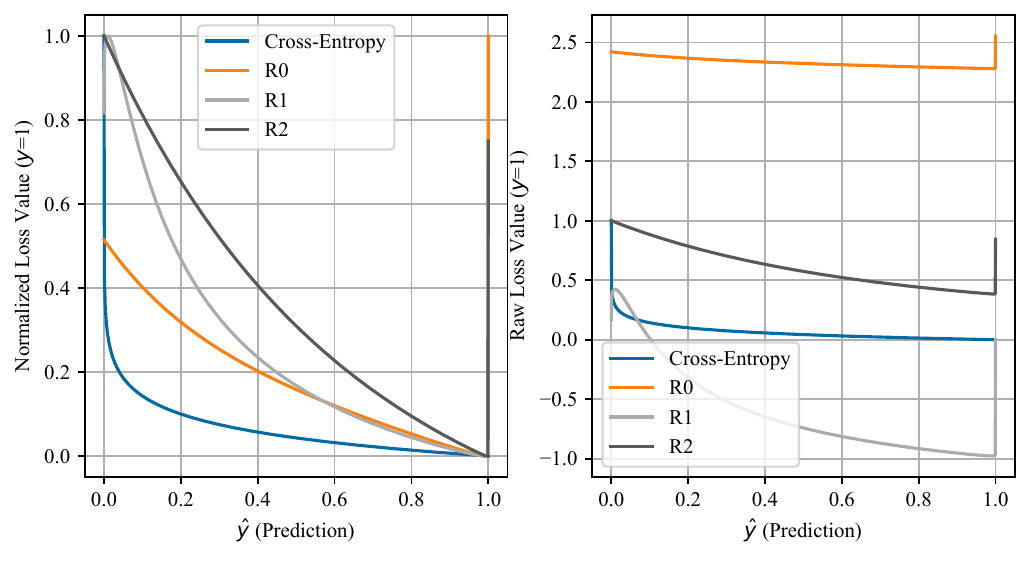}}
\caption{Binary Phenotype of the final RandAug losses against cross-entropy for both normalized and raw loss values.} 
\label{fig:ra_losses}
\end{center}
\vskip -0.1in
\end{figure}

\begin{figure}[htb]
\vskip 0.1in
\begin{center}
\centerline{\includegraphics[width=0.5\columnwidth]{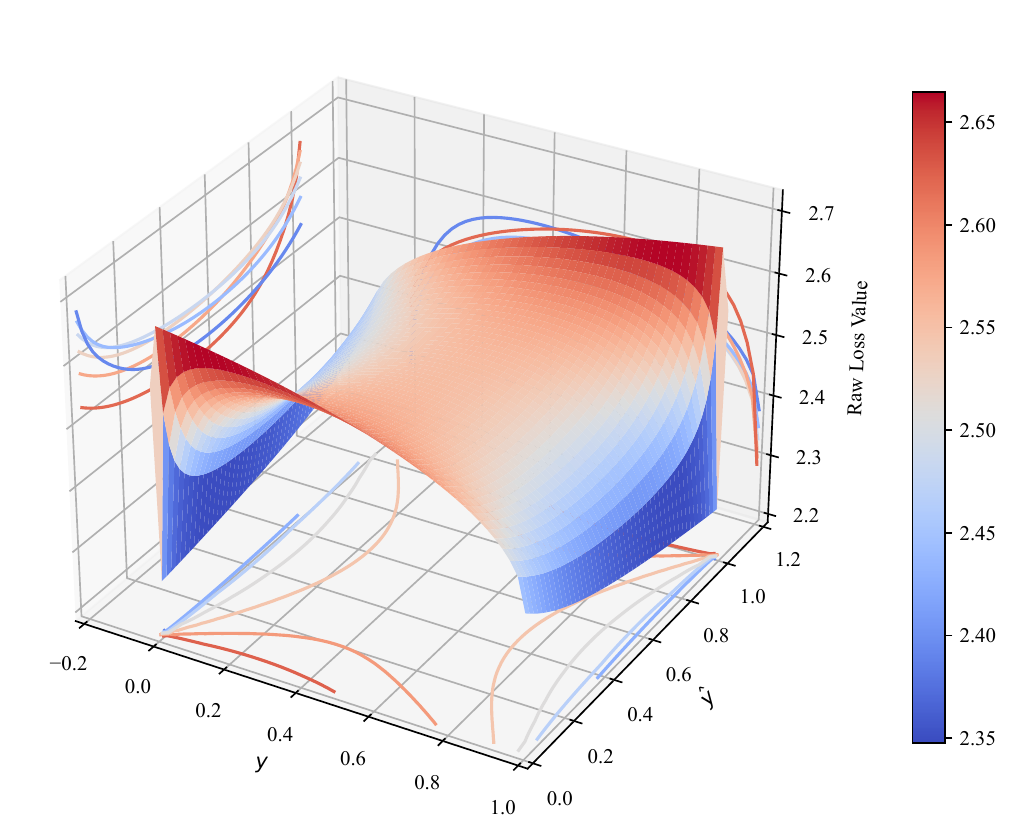}}
\caption{3D plot of the raw phenotype of the $R0$ loss.} 
\label{fig:r0_3D}
\end{center}
\vskip -0.1in
\end{figure}

\begin{figure}[htb]
\vskip 0.1in
\begin{center}
\centerline{\includegraphics[width=0.5\columnwidth]{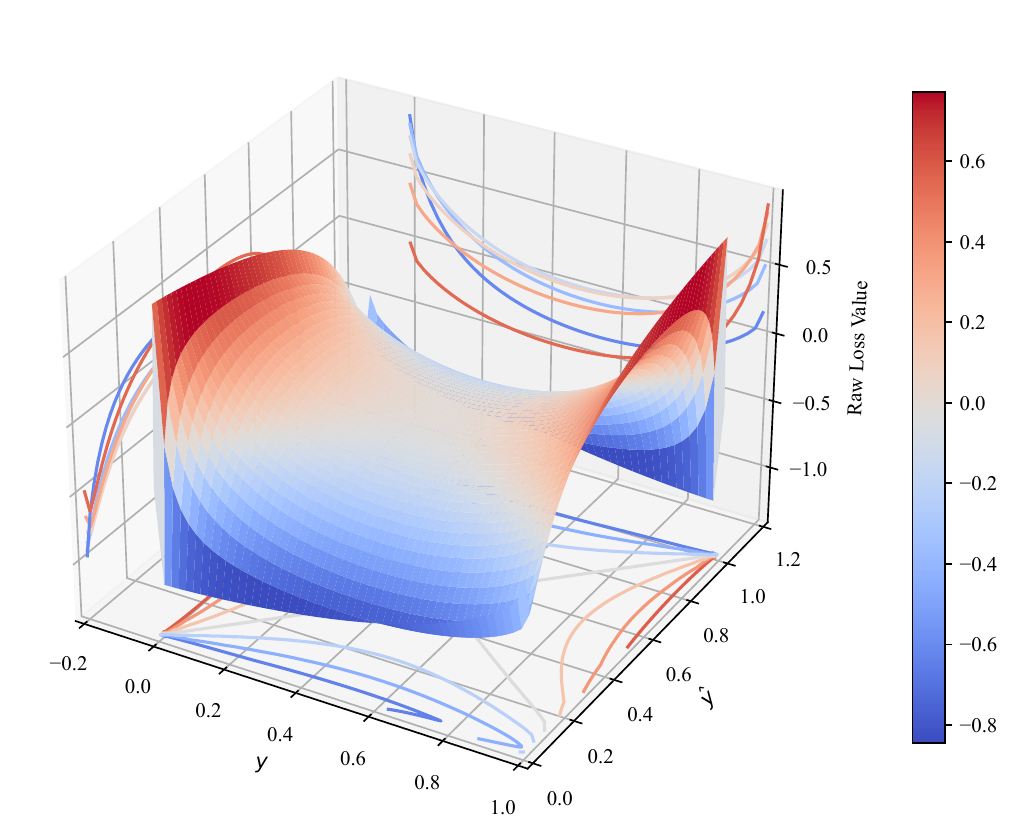}}
\caption{3D plot of the raw phenotype of the $R1$ loss.} 
\label{fig:r1_3D}
\end{center}
\vskip -0.1in
\end{figure}

\begin{figure}[htb]
\vskip 0.1in
\begin{center}
\centerline{\includegraphics[width=0.5\columnwidth]{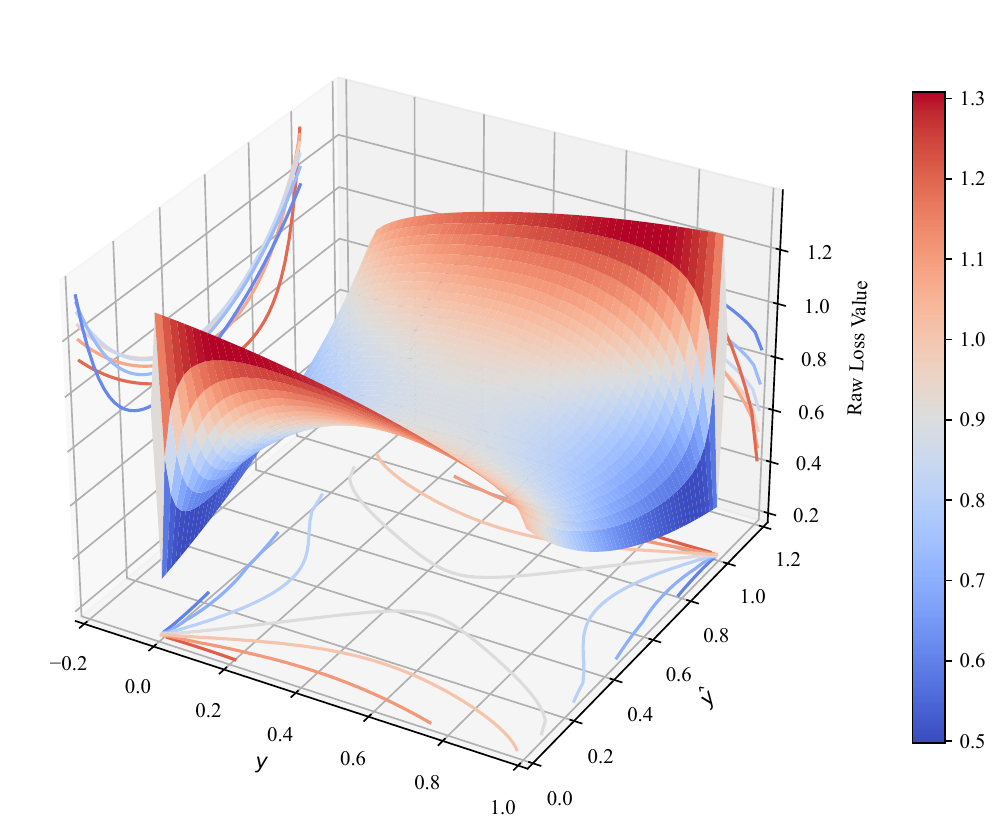}}
\caption{3D plot of the raw phenotype of the $R2$ loss.} 
\label{fig:r2_3D}
\end{center}
\vskip -0.1in
\end{figure}

The normalized and raw binary phenotypes of the final mixup losses are shown in Figure~\ref{fig:ra_losses}. The non-normalized three-dimensional representations for $R0$, $R1$, and $R2$ are shown in Figures~\ref{fig:r0_3D}, \ref{fig:r1_3D}, and \ref{fig:r2_3D}. As one can see from the binary phenotypes, the final RandAug functions host a variety of differences within themselves. Mainly, $R0$ is the only loss function discovered where its maximum loss value does not occur when $\hat y=0$ for $y=1$, but actually when $\hat y=1$. Meaning, $R0$ gives the greatest penalty to when the model's output directly matches the target. It seems that this level of regularization is too much for deep learning models as $R0$ performs extremely poorly on CIFAR-100. 

\subsection{All Losses}

\begin{figure}[htb]
\vskip 0.1in
\begin{center}
\centerline{\includegraphics[width=0.5\columnwidth]{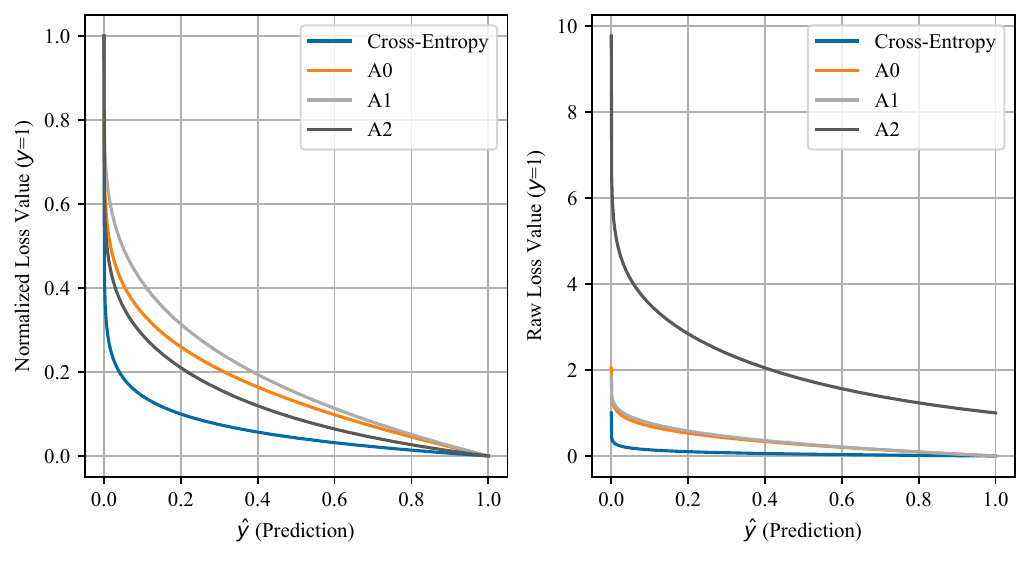}}
\caption{Binary Phenotype of the final All losses against cross-entropy for both normalized and raw loss values.} 
\label{fig:all_losses}
\end{center}
\vskip -0.1in
\end{figure}

\begin{figure}[htb]
\vskip 0.1in
\begin{center}
\centerline{\includegraphics[width=0.5\columnwidth]{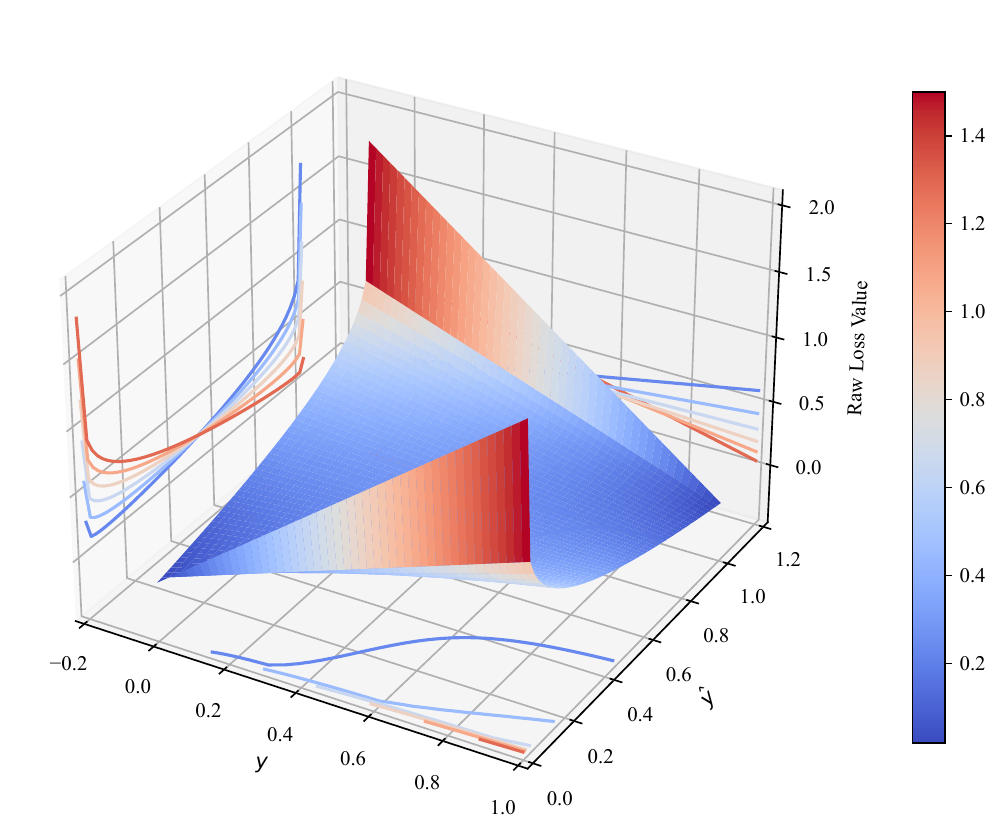}}
\caption{3D plot of the raw phenotype of the $A0$ loss.} 
\label{fig:a0_3D}
\end{center}
\vskip -0.1in
\end{figure}

\begin{figure}[htb]
\vskip 0.1in
\begin{center}
\centerline{\includegraphics[width=0.5\columnwidth]{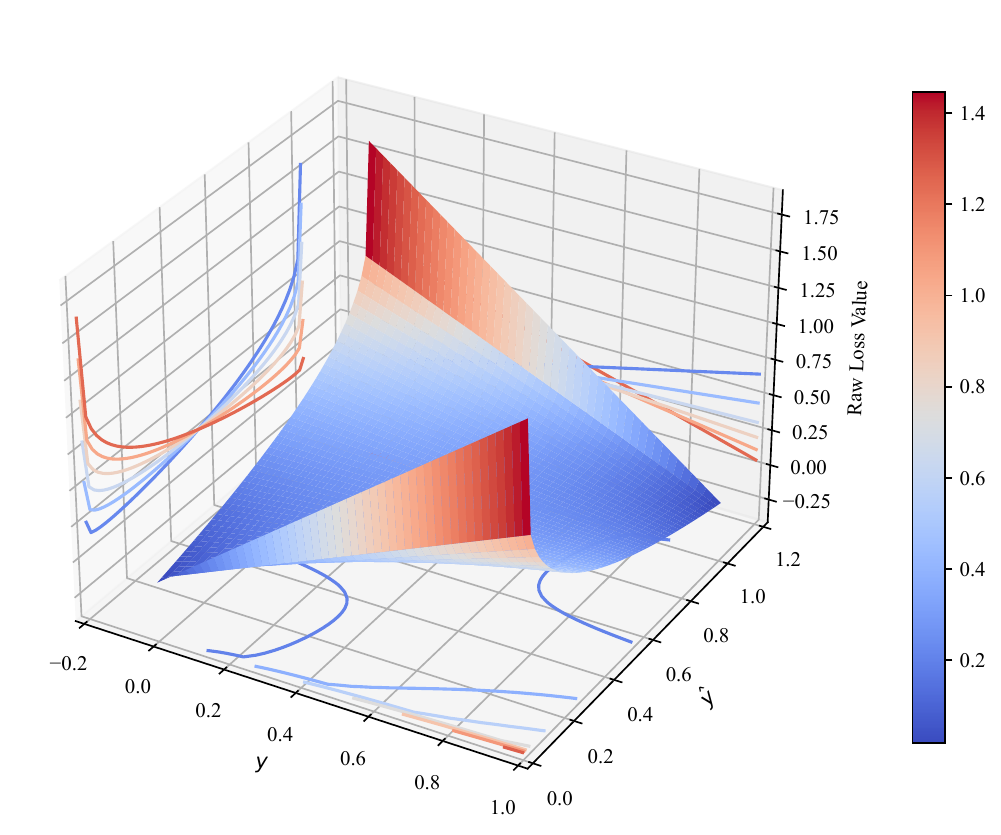}}
\caption{3D plot of the raw phenotype of the $A1$ loss.} 
\label{fig:a1_3D}
\end{center}
\vskip -0.1in
\end{figure}

The normalized and raw binary phenotypes of the final mixup losses are shown in Figure~\ref{fig:all_losses}. The non-normalized three-dimensional representations for $A0$, $A1$, and $A2$ are shown in Figures~\ref{fig:a0_3D} and \ref{fig:a1_3D}. As one can see from the binary and three-dimensional phenotypes, the three final all functions are extremely similar in shape. The main difference is $A2$ has larger raw loss values. Empirically, these results seem to support the larger raw loss values as $A2$ consistently outperformed both $A0$ and $A1$.

\end{document}